\documentclass[graybox]{svmult}

\usepackage{type1cm}        % activate if the above 3 fonts are
\usepackage{makeidx}         % allows index generation
\usepackage{graphicx}
\usepackage[caption=false]{subfig}
\usepackage{wrapfig}
\usepackage{multicol}        % used for the two-column index
\usepackage[bottom]{footmisc}% places footnotes at page bottom

\usepackage{amsmath}
\usepackage{newtxtext}       % 
\usepackage{newtxmath}       % selects Times Roman as basic font
\usepackage{mathtools}

\usepackage{pgfplots}
\usepackage{soul}

\usepackage{tikz}
\usetikzlibrary{calc,positioning,fit} 
\graphicspath{ {../} }%
\usepackage{algorithm} % for pseudocode
\usepackage{algpseudocode} % for pseudocode

\usepackage{dirtytalk} % for both-sided quotation marks " "

\usepackage{comment} % for comments

\usepackage{gensymb} % for degree symbol

\usepackage{hyperref} % for including hyperlinks
\hypersetup{
    colorlinks=true,
    linkcolor=blue,
    filecolor=magenta,      
    urlcolor=blue,
    pdftitle={Safe Gap-based Planning in Dynamic Settings},
    pdfpagemode=FullScreen,
}

\usepackage{multirow}  
\usepackage{array} 
\usepackage{tabularx}

\usepackage[rightcaption]{sidecap}

\sidecaptionvpos{figure}{c}

\usepackage{ragged2e} 

\newcounter{refnavchap}[section]

\DeclareMathOperator{\dotprod}{dot}

\makeindex             % used for the subject index
\begin{document}

\refstepcounter{refnavchap}
\label{ch:egonav}

\title*{Safe Gap-based Planning in Dynamic Settings} \label{ch:dgap}
\author{Max Asselmeier, Abdel Zaro, Dhruv Ahuja, Ye Zhao, and Patricio A. Vela}
\institute{M. Asselmeier $\cdot$ A. Zaro $\cdot$ Y. Zhao $\cdot$ P. A. Vela \at Institute for Robotics and Intelligent Machines, Georgia Institute of Technology, Atlanta, GA 30308. \email{mass@gatech.edu, azaro3@gatech.edu, yzhao301@gatech.edu, pvela@gatech.edu}
\and D. Ahuja \at School of Electrical and Computer Engineering, Georgia Institute of Technology, Atlanta, GA 30308. \email{dahuja8@gatech.edu}
\and This work supported in part by NSF Award \#2235944. The work of Max Asselmeier is supported by the NSF Graduate Research Fellowship under Grant No. DGE-2039655. Any opinion, findings, and conclusions or recommendations expressed in this material are those of the authors(s) and do not necessarily reflect the views of the National Science Foundation.
}
\maketitle

\abstract*{This chapter extends the family of perception-informed gap-based local planners to dynamic environments. Existing perception-informed local planners that operate in dynamic environments often rely on emergent or empirical robustness for collision avoidance as opposed to performing formal analysis of dynamic obstacles. This proposed planner, \textit{dynamic gap}, explicitly addresses dynamic obstacles through several steps in the planning pipeline. First, polar regions of free space known as gaps are tracked and their dynamics are estimated in order to understand how the local environment evolves over time. Then, at planning time, gaps are propagated into the future through novel gap propagation algorithms to understand what regions are feasible for passage. Lastly, pursuit guidance theory is leveraged to generate local trajectories that are provably collision-free under ideal conditions. Additionally, obstacle-centric \textit{ungap} processing is performed in situations where no gaps exist to robustify the overall planning framework. A set of gap-based planners are benchmarked against a series of classical and learned motion planners in dynamic environments, and dynamic gap is shown to outperform all other baselines in all environments. Furthermore, dynamic gap is deployed on a TurtleBot2 platform in several real-world experiments to validate collision avoidance behaviors.}

\abstract{This chapter extends the family of perception-informed gap-based local planners to dynamic environments. Existing perception-informed local planners that operate in dynamic environments often rely on emergent or empirical robustness for collision avoidance as opposed to performing formal analysis of dynamic obstacles. This proposed planner, \textit{dynamic gap}, explicitly addresses dynamic obstacles through several steps in the planning pipeline. First, polar regions of free space known as gaps are tracked and their dynamics are estimated in order to understand how the local environment evolves over time. Then, at planning time, gaps are propagated into the future through novel gap propagation algorithms to understand what regions are feasible for passage. Lastly, pursuit guidance theory is leveraged to generate local trajectories that are provably collision-free under ideal conditions. Additionally, obstacle-centric \textit{ungap} processing is performed in situations where no gaps exist to robustify the overall planning framework. A set of gap-based planners are benchmarked against a series of classical and learned motion planners in dynamic environments, and dynamic gap is shown to outperform all other baselines in all environments. Furthermore, dynamic gap is deployed on a TurtleBot2 platform in several real-world experiments to validate collision avoidance behaviors.}

% First version from ICRA
% This paper extends the family of gap-based local planners to unknown dynamic environments through generating provably collision-free properties for hierarchical navigation systems. Existing perception-informed local planners that operate in dynamic environments rely on emergent or empirical robustness for collision avoidance as opposed to performing formal analysis of dynamic obstacles. In addition to this, the obstacle tracking that is performed in these existent planners is often achieved with respect to a global inertial frame, subjecting such tracking estimates to transformation errors from odometry drift. The proposed local planner,  \textit{dynamic gap}, shifts the tracking paradigm to modeling how the free space, represented as gaps, evolves over time. Gap crossing and closing conditions are developed to aid in determining the feasibility of passage through gaps, and a breadth of simulation benchmarking is performed against other navigation planners in the literature where the proposed dynamic gap planner achieves the highest success rate out of all planners tested in all environments.

\section*{Acronyms}
\begin{description}[ABCDEF]
\item[A3C]{Asynchronous Advantage Actor-Critic}
\item[CADRL]{Collision Avoidance with Deep Reinforcement Learning}
\item[CoHAN]{Co-operative Human-aware Navigation}
\item[DRL-VO]{Deep Reinforcement Learning - Velocity Obstacles}
\item[DGap]{Dynamic Gap}
\item[DWA]{Dynamic Window Approach}
\item[FOV]{Field of View}
\item[GBP]{Gap-based Planner}
\item[HA]{Hungarian Algorithm}
\item[ICP]{Iterative Closest Point}
\item[IQA]{Iterative Quaternion Averaging}
\item[KF]{Kalman Filter}
\item[LfLH]{Learning from Learned Hallucination}
\item[LSTM]{Long Short-Term Memory}
\item[MPC]{Model Predictive Control}
\item[ORCA]{Optimal Reciprocal Collision Avoidance}
\item[PN]{Parallel Navigation}
\item[PGap]{Potential Gap}
\item[PRM]{Probabilistic Roadmap}
\item[PPO]{Proximal Policy Optimization}
\item[PP]{Pure Pursuit}
\item[RRT]{Rapidly-exploring Random Trees}
\item[RVO]{Reciprocal Velocity Obstacles}
\item[RA]{Rectangular Assignment}
\item[RLCA]{Reinforcement Learning for Collision Avoidance}
\item[ROS]{Robot Operating System}
\item[TEB]{Timed Elastic Bands}
\item[VO]{Velocity Obstacles}
\end{description}

\section{Introduction}
\label{sec:introduction}

% P1: SOTA for gap-based planning
The family of gap-based planners (GBP) offers a free space-based approach to real-time navigation in previously unseen environments. GBPs seek to leverage the environmental affordances \cite{gibson_ecological_2014} available to the robot and restrict all planning efforts to these free regions, or \textit{gaps}, in the environment. In the literature, this family of planners has shown great promise due to their ability to provide safety guarantees \cite{xu_potential_2021} under ideal assumptions including full field-of-view (FOV) sensing, first-order holonomic dynamics, and point-mass geometry. Beyond these ideal conditions, GBPs have still provided impressive navigation performance given restricted FOVs \cite{xu_potential_2021}, noncircular robot geometries \cite{feng_gpf-bg_2023}, and nonholonomic dynamics \cite{feng_safer_2023}.

% P2: Research deficit for gap-based planners: dynamic environments
Despite this success, GBPs have only been extended to handling dynamic obstacle avoidance very recently \cite{chen_safe_2022, contarli_predictive_2024}, a challenge that accurately reflects the previously unseen and constantly changing environments found in real life. Typically, local motion planners are designed with static environments in mind. Then, when the time comes to deploy these planners in dynamic real-world settings, the planners are often run with the hopes of high enough planning rates enabling sufficient reactive collision avoidance. However, relying on these emergent or empirical behaviors limits planner performance. It is more meaningful to design a planner with dynamic settings in mind, but this often requires each and every step of the planning pipeline to be revisited to account for these relaxed assumptions. In static settings, the currently available free space will remain as free space for all time. Naturally, the same applies for obstacle space. In dynamic settings, free space can turn into obstacle space, and vice-versa. If a motion planner is to fully exploit its environmental affordances, then both of these possibilities must explicitly be accounted for. 

\begin{SCfigure}
    \includegraphics[width=0.50\linewidth]{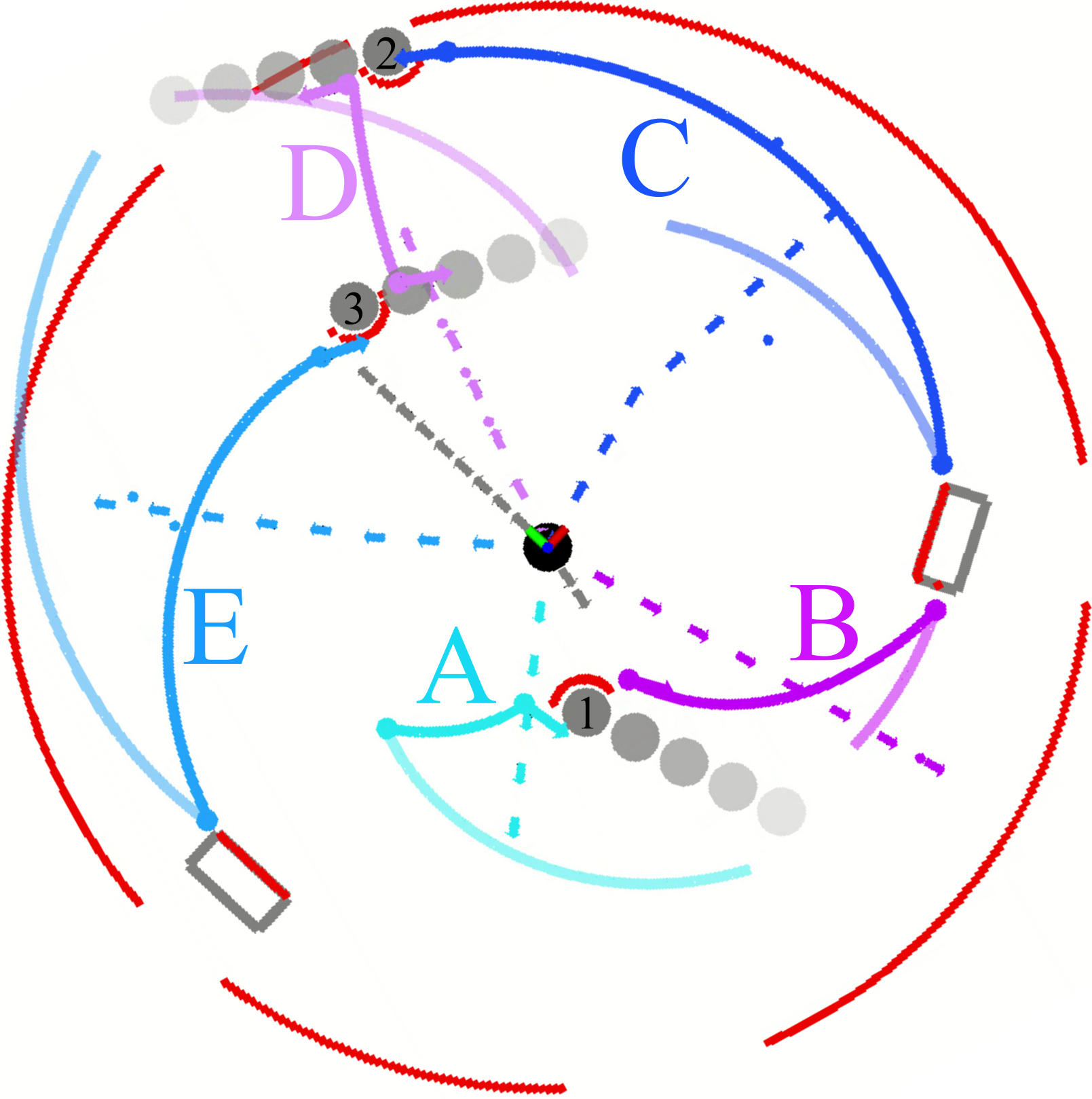}
    \caption{Visualization of the dynamic gap planner. The black disk is the ego-robot and gray disks are dynamic agents. Transparent gray disks are future agent positions. Bold colored arcs labeled A-E are the instantaneous set of gaps. Transparent arcs are the predicted gaps obtained by propagating the gap dynamics models, shown as arrows, forward. Dashed lines are the candidate trajectories and the gap goals are colored points within the gap spans. The pink gap tube labeled D, which is comprised of all present and future pink gap arcs, is predicted to close as agents 2 and 3 cross each other. However, this gap will reopen, represented by the transparent pink arc. The gap tube trajectory plans up to the gap closure, idles in place while the gap is closed, and then continues through the reopened gap. Ungap trajectories, represented as gray dashed lines, are synthesized in the receding ungaps formed by agents 1 and 2.}
    \label{fig:dgap_vis}
\end{SCfigure}

\begin{comment}
\begin{figure}[t]
  \centering
  \includegraphics[width=0.65\linewidth]{figures/vision.png}
  \caption{Visualization of gaps and trajectories generated by dynamic gap. The central black disk depicts the ego-robot while the gray disks depict dynamic agents. The solid gray disks represent the current positions of the agents and the increasingly transparent gray disks represent the future positions of the agents. The bold colored arcs labeled A-E are the instantaneous set of gaps that are used for planning. The transparent arcs are the predicted gaps obtained by propagating the gap dynamics models, shown as arrows, forward in time. Dashed lines are the candidate trajectories synthesized towards the gap goals which are represented as colored points within the gap spans. The pink gap tube with the label D, which is comprised of all present and future pink gap arcs, is predicted to close as the agents labeled $2$ and $3$ cross each other. However, this gap will reopen later on, represented by the transparent pink arc. The gap tube trajectory plans up to the point of the gap closure, idles in place while the gap is closed, and then continues through the reopened gap towards the further gap goal. Ungap trajectories, represented as gray candidate trajectories, are synthesized in the receding ungaps formed by agents 1 and 2 as well. }
  \label{fig:dgap_vis}
\end{figure}
\end{comment}

% P3: Introduce work, aims, and desired takeaways
To date, GBPs designed for dynamic settings \cite{chen_safe_2022, contarli_predictive_2024} have not been perception-informed, instead opting to rely on ground truth state information regarding other agents in the local environment. The intention of this proposed work is to design a perception-informed GBP for dynamic settings in order to explicitly account for both sensor data artifacts and potentially evolving environments. In \cite{asselmeier_dynamic_2025}, the authors introduced an initial version of the \textit{dynamic gap} planner as a proof of concept. In this book chapter, the dynamic gap planner is documented in greater detail to further elucidate each intermediate step in the planning pipeline. The planner itself is extended by relaxing two limiting assumptions: (1) the \say{isolated gap} assumption in which gaps were previously not allowed to interact during gap propagation, and (2) the \say{unsafe ungap} assumption in which the polar spaces between gaps caused by obstacles were deemed entirely uninhabitable and unsafe to plan within. In summary, the core contributions of the proposed work are as follows.
\begin{enumerate}   
    \item Proposing an alternative tracking paradigm which revolves around the tracking and prediction of free space
    \item Adapting geometric and kinematic rules from guidance laws to the realm of GBPs to aid in dynamic gap propagation and feasibility analysis
    \item Performing complex gap propagation analysis to enable planning through interrupted and switched gaps
    \item Providing a proof of collision-free dynamic gap passage under ideal conditions along with supplementary safety modules for non-ideal conditions
    \item Benchmarking against state-of-the-art local planners in the Arena-Rosnav simulation environment
    \item Validating planner performance onboard a TurtleBot2 platform in dynamic real-world environments
\end{enumerate}

\begin{figure}[t!] \label{fig:info_flow}
    \centering
    \begin{tikzpicture}
          \hspace*{-0.5in}
        \node[] at ($(0, 0)$){{\scalebox{0.9} {\input{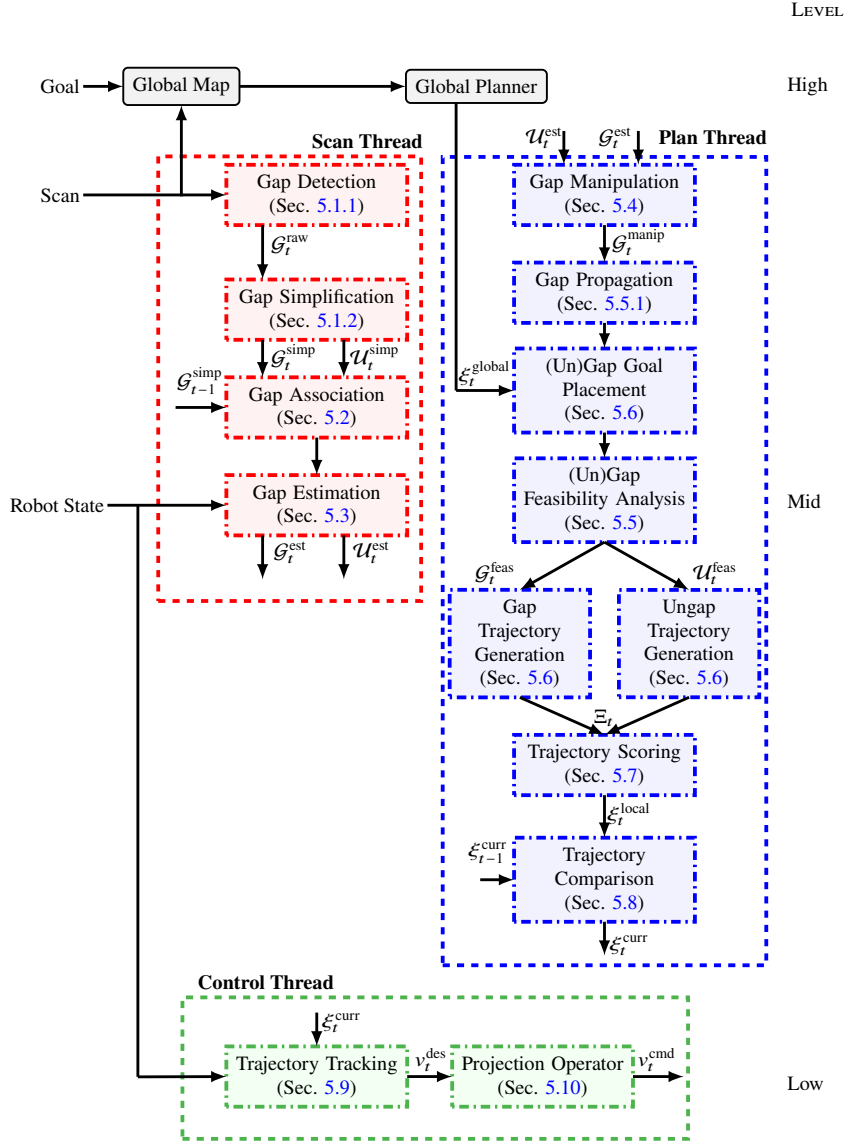}}}};
    \end{tikzpicture}
    \caption{Workflow for navigation framework. Within this framework, there are three active threads: scan, plan, and control. The scan thread (red) is run at the rate of the incoming laser scan, which is $25$ Hz in simulation and on hardware. This thread acts on an incoming laser scan $\mathcal{L}$ and outputs a set of estimated gaps $\mathcal{G}_t^{\rm est}$ and ungaps $\mathcal{U}_t^{\rm est}$. The plan thread (blue) then synthesizes a set of candidate local trajectories $\Xi_t$ through these estimated gaps and ungaps. Finally, the control thread acts to track the selected local trajectory $\xi^{\rm curr}_t$. Both the plan and control thread are bound to run together at the prescribed planning rate which is $5$ Hz in simulation and on hardware.} 
    % \vspace{-0.25in}
\end{figure}

A visual snapshot of active components during local planning for dynamic gap is portrayed in Figure \ref{fig:dgap_vis}. This planner is open-sourced\footnote{\href{https://github.com/ivaROS/DynamicGap}{https://github.com/ivaROS/DynamicGap}} in Arena-Rosnav \cite{kastner_arena-rosnav_2021, kastner_arena-bench_2022, kastner_arena-rosnav_2023, kastner_arena_2024} and available to test. The information flow for the planner can be seen in Figure \ref{fig:info_flow}. In this chapter, the task of navigation in dynamic environments is defined (Section \ref{sec:motion_planning_in_dynamic_environments}) and common planning pipelines (\ref{sec:planning_considerations}) and approaches (\ref{sec:planning_approaches}) are surveyed and discussed. Social navigation considerations that become important in dynamic real-world settings are explained in Section \ref{sec:social_navigation}. The paradigm of Planning in the Perception Space (PiPS) that GBPs fall under is discussed in Section \ref{sec:planning_in_the_perception_space}. Preliminary concepts for GBPs in dynamic environments are defined (Section \ref{sec:preliminaries}) before the proposed local planner is detailed (Section \ref{sec:dynamic_gap}). Then, experimental results in Section \ref{sec:experimental_results} validate theoretical claims of collision avoidance (\ref{sec:experiment_1}), illuminate planning performance in more contrived setups (\ref{sec:experiment_2}), and contextualize dynamic gap planning performance among state-of-the-art planners (\ref{sec:simulation_benchmarking}). Analysis of the computational workload of the dynamic gap planner (\ref{sec:timing}) demonstrates the efficiency of the PiPS framework, social compliance results (\ref{sec:social}) provide insight into the behavior of dynamic gap around pedestrians in simulation, and, lastly, hardware experiments (\ref{sec:hardware_testing}) validate this proposed planning framework in the real world.

\section{Navigation in Dynamic Environments}
\label{sec:motion_planning_in_dynamic_environments}
% need more citations for when we introduce data clustering/association/estimation steps.
Mobile robot planning concerns the generation of a sequence of motion commands for a mobile robot platform to maneuver from an arbitrary start position to an arbitrary goal position. If this environment is entirely known and unchanging, then this task is purely a motion planning problem. If the environment is previously unknown, partially observable, and potentially even changing over time, then this task becomes a navigation problem. For this navigation problem, the robot must also possess some means to reveal the environment structure online through visual sensing. This mobile robot may also be subject to geometric constraints due to its size and shape, kinematics constraints due to its composition and linkages, or dynamics constraints due to its actuation capabilities. In the context of this chapter, the robot must reach its goal position (1) under a predefined time limit and (2) without colliding into any static or dynamic artifacts in the environment in order for the task to be deemed a success.

\subsection{Planning Considerations} \label{sec:planning_considerations}
When navigating in a dynamic environment, a handful of additional considerations must be made when compared to static environments: \textit{data clustering}, \textit{data association}, and \textit{data estimation}. First, a navigation framework should have some way to \textit{cluster} or abstract raw sensor data into meaningful primitives that enable local planners to efficiently avoid environment collisions. These primitives can take many forms, including point clusters \cite{wang_model-free_2015, vaskov_towards_2019}, disks \cite{vasilopoulos_reactive_2020}, polygonal geometries \cite{rosmann_timed-elastic-bands_2015, rosmann_track_2018}, or semantic labels \cite{salzmann_trajectron_2021}. Second, a navigation framework must find and \textit{associate} pairings of these primitives between consecutive time steps in order to retain a history or memory of primitives. Depending on the format of the primitive, registration algorithms such as Iterative Closest Point (ICP) \cite{besl_method_1992} or Iterative Quaternion Averaging (IQA) \cite{mishra_efficient_2024} can provide pairwise assignments. Other approaches treat this step as a Rectangular Assignment (RA) problem and solve via the Hungarian Algorithm (HA) \cite{kuhn_hungarian_1955}. Lastly, a navigation framework must \textit{estimate} the unobservable portion of the state of the dynamic obstacles. This involves updating a set of prediction models that could be variants of Kalman filters (KF) \cite{wang_model-free_2015, guizilini_dynamic_2019, xu_real-time_2023}, Gaussian processes \cite{senanayake_bayesian_2017}, factor graphs \cite{poschmann_factor_2020}, or neural networks \cite{salzmann_trajectron_2021}.

Once all of these steps have been taken, prediction models can be factored into planning decisions through capturing them in cost functions or constraints in Model Predictive Control (MPC) formulations \cite{gaertner_collision-free_2021, rosmann_timed-elastic-bands_2015}, or embedding them in occupancy maps that can be searched over for paths \cite{senanayake_spatio-temporal_2016, guizilini_dynamic_2019}. The methods cited here assign prediction models to individual obstacles and represent these prediction models with respect to a fixed frame. Performing obstacle tracking in the egocentric frame allows the local planner to leverage the benefits of operating in the perception space which will be discussed in Section \ref{sec:planning_in_the_perception_space}.

% The task of navigating through dynamic environments can be approached at various levels. For one, there can be a focus on the tracking and prediction of dynamic obstacles. Such approaches can be categorized into two approaches: model-based and motion-based. Model-based methods involve classifying obstacles from a predefined set of categories such as pedestrians or automobiles. This usually involves visual inferencing from a learned classifier \cite{salzmann_trajectron_2021}. Model-based methods are advantageous in that prediction models can be catered towards specific behaviors or classes \cite{zhao_qualitative_1998}, and they can identify objects in the scene that may not currently be moving but could move in the future. However, these approaches usually require a high level of confidence in the object classifier. 

% Motion-based methods involve detecting dynamic obstacles from sensor-derived motion cues. Gap-based planners fall into this category. These methods are advantageous in that they will capture any moving artifact in the environment, but they are disadvantageous in that they model all artifacts identically. To make obstacle predictions, motion-based methods perform data clustering, data association, and data estimation. 

\subsection{Planning Approaches} \label{sec:planning_approaches}
Within the mobile robotics research community, navigation is a well-studied problem with a myriad of approaches. Many classical algorithms were originally designed for static environments and later saw extensions to explicitly account for dynamic environments. Graph search-based methods such as A* \cite{hart_formal_1968} can discretize the continuous configuration space of a robot and search over it for a collision-free path to the goal. Later, the methods D* \cite{stentz_dalgorithm_nodate} and D* Lite \cite{s_koenig_and_m_likhachev_d_nodate} were designed to handle the dynamic environment case. Sampling-based methods such as Probabilistic Roadmaps (PRM) \cite{kavraki_probabilistic_1996}, Rapidly-Exploring Random Trees (RRT) \cite{lavalle_rapidly-exploring_1998}, and their asymptotically optimal extensions PRM* and RRT* \cite{karaman_sampling-based_2011} can be viewed as extensions of graph-based methods that opt to randomly sample the continuous configuration space instead of discretization. This family of planners has also seen extensions to dynamic environments \cite{huppi_t-prm_2022, zucker_multipartite_2007}. Some planning algorithms such as the Dynamic Window Approach (DWA) \cite{fox_dynamic_1997} choose to sample in the control space as opposed to the configuration space and then roll out these sampled control inputs. Similarly, dynamic versions of DWA have been developed \cite{missura_predictive_2019}. More recently, trajectory optimization has proven to be a potent form of motion planning that can account for kinodynamical constraints while providing optimal trajectories. Methods including Timed Elastic Bands (TEB) \cite{rosmann_timed-elastic-bands_2015} formulate local navigation tasks as optimal control problems which are also general enough to handle dynamic environments \cite{rosmann_track_2018}. Some approaches such as velocity obstacles (VOs) \cite{fiorini_motion_1993, fiorini_motion_1998}, reciprocal velocity obstacles (RVOs) \cite{van_den_berg_reciprocal_2008}, and optimal reciprocal collision avoidance (ORCA) \cite{siciliano_reciprocal_2011} were initially designed with dynamic obstacles in mind. Methods from control theory have also been leveraged for effective collision avoidance strategies including Hamilton-Jacobi-Bellman reachability-based approaches \cite{kousik_bridging_2020, vaskov_guaranteed_2019} and receding horizon-based methods \cite{brito_model_2020, vahs_forward_2024, heuer_proactive_2023, de_groot_topology-driven_2025}. Lower-level safety filters such as control barrier functions \cite{lafmejani_nmpc-lbf_2022, long_safe_2024, so_how_2023} and backup controllers \cite{agrawal_gatekeeper_2023} can also modify control inputs to ensure safety during online deployment. 

As of late, data-driven approaches to local navigation and collision avoidance have also received tremendous amounts of attention. Most of these learned planners are based on deep reinforcement learning \cite{lutzow_density_2023, long_towards_2018, everett_collision_2021, xu_applr_2020, xie_drl-vo_2023, wang_agile_2021} due to the availability of physics simulations and self-supervised trial-and-error-based data collection methods. Such planners have seen significant use due to their apparent generalizability and scalability. However, these methods have yet to see widespread integration onto hardware platforms due to their lack of safety and interpretability. Extensive comparisons of classical and learned methods are still difficult to find \cite{xiao_motion_2022, xu_benchmarking_2023}, and one of the motivations of this chapter is to provide a detailed benchmarking comparison of several state-of-the-art planners, both learned and classical. One of the particularly exciting facets of the proliferation of robotics research is the rise in publically available competitions and challenges to stimulate progress. Both the Benchmark Autonomous Robot Navigation challenge \cite{xiao_autonomous_2024} and DynaBARN \cite{nair_dynabarn_2022} challenges have been excellent places to locate and evaluate state-of-the-art local navigation planners.
% (BARN)

\subsection{Social Navigation}
\label{sec:social_navigation}
Under the context of navigation, planning in dynamic environments commonly means planning in the presence of humans. Social navigation refers to a robot’s ability to navigate in a manner that goes beyond geometric efficiency, such as minimizing path length, to also account for social considerations, including human comfort, and norms. A socially compliant robot must be capable of recognizing people and prioritizing their safety accordingly. Beyond collision avoidance, the planner's behavior should aim to minimize discomfort, reduce confusion, and maintain predictability in its movements. Socially-aware navigation also involves the ability to convey intent, either through movement patterns or explicit signals, so that nearby humans can understand and anticipate the robot's actions. In situations where human and robot objectives may conflict, a socially compliant robot may even compromise its own efficiency to resolve the situation in a way that aligns with human expectations \cite{Singamaneni2024Feb}. % Ultimately, social navigation expands the objective of robot motion planning beyond reaching a goal.

 Failure to navigate in a socially compliant manner can cause discomfort and even increase the risk of collisions, especially in situations where human behavior is unpredictable.  Social considerations are particularly important as robots become increasingly integrated into everyday social environments, supporting applications such as food delivery, healthcare settings, and office spaces. However, human behavior varies widely across contexts, individuals may be distracted by their phones, uncooperative towards robots, or hurried by their own tasks, introducing patterns that the robot may not be familiar with. This variability makes it challenging to develop navigation approaches that generalize effectively across real-world environments.

Numerous approaches have been proposed to address this. Classical geometric methods with hand-engineered cost functions and heuristics dominate industrial deployment due to their reliability and ease of software integration. Data-driven approaches, on the other hand, are able to address many of the shortcomings of classical based approaches, such as capturing nuanced human behaviors, predicting trajectories and incorporating social norms; however, they lack the same safety guarantees and explainability and have higher integration overhead \cite{Amir}. 

In order to assess a planner's performance in a social setting, various metrics such as the velocity or time spent in personal space can be used, where personal space is often defined as some radius around humans \cite{arena30}. 

% \subsection{Motion Planning in Dynamic Environments}

\section{Planning in the Perception Space}
\label{sec:planning_in_the_perception_space}
Most motion planners \cite{lavalle_rapidly-exploring_1998, rosmann_timed-elastic-bands_2015, connell_dynamic_2017} opt to plan using Cartesian world-frame environment representations such as occupancy maps or voxel grids. These approaches contrast the perception space approach to planning which involves keeping sensory input in its raw egocentric form to take advantage of the computational benefits that come with foregoing intensive data processing. Applying the PiPS framework means that all local planning steps downstream must then be cast as egocentric decision-making processes. 

Prior work from the authors has developed a hierarchical perception space navigation stack fit with efficient collision checking \cite{smith_pips_2017}, egocentric environment representations \cite{smith_real-time_2020}, and local path planning \cite{smith_egoteb_2020, xu_potential_2021}. 

GBPs \cite{sezer_novel_2012, mujahed_admissible_2018, mujahed_safe_2013, mujahed_new_2016, feng_gpf-bg_2023, mujahed_tangential_2013, mujahed_admissible_2017, mujahad_closest_2010, ullah_fnug_2022, demir_improved_2017, contarli_predictive_2024, feng_safer_2023, chen_safe_2022, groot_topology-driven_2024} are a form of perception space-based planners that detect regions of collision-free space defined by leading or trailing edges of obstacles. This can be viewed as an alternative way of discretizing the environment, here in the egocentric polar space as opposed to discretizing in the Cartesian space for common methods such as occupancy maps \cite{ocallaghan_gaussian_2012, ramos_hilbert_2016, missura_polygonal_2020}. Some attention has been given to gaps in dynamic environments \cite{chen_safe_2022, contarli_predictive_2024}, but these methods do not develop their theory through a perception-informed approach, instead opting to use ground truth agent pose information. In the proposed work, explicit attention is paid towards how the dynamics of gaps must be ascertained from scan data in order to understand how the local gaps evolve over time.

\section{Preliminaries}
\label{sec:preliminaries}
First, the primitive environmental affordances that the proposed planner will exploit for planning are defined and visualized.

\subsection{Gap}
\textit{Definition (Gap):} A polar region of free space in $\mathbb{R}^2$ characterized by the area swept out between $\mathbf{p}_{l/e}$ and $\mathbf{p}_{r/e}$ where the subscripts $l/e$ and $r/e$ represent relative measurements between the left and right gap points, respectively, and the ego-robot. \\
Gap points are also assigned velocities $\mathbf{v}_{l/e}$ and $\mathbf{v}_{r/e}$. Together, these form the left and right gap point states $\mathbf{X}_{l/e} = \begin{bmatrix} \mathbf{p}_{l/e} & \mathbf{v}_{l/e} \end{bmatrix}^T$ and $\mathbf{X}_{r/e} = \begin{bmatrix} \mathbf{p}_{r/e} & \mathbf{v}_{r/e} \end{bmatrix}^T$.

%  = \begin{bmatrix} r_{l/e} \cos(\beta_{l/e}) & r_{l/e} \sin(\beta_{l/e}) \end{bmatrix}^T

%  = \begin{bmatrix} r_{r/e} \cos(\beta_{r/e}) & r_{r/e} \sin(\beta_{r/e}) \end{bmatrix}^T

% $\dot{\mathbf{p}}_{l/e} = \begin{bmatrix} \dot{x}_{l/e} & \dot{y}_{l/e} \end{bmatrix}^T$ and $\dot{\mathbf{p}}_{r/e} = \begin{bmatrix} \dot{x}_{r/e} & \dot{y}_{r/e} \end{bmatrix}^T$

% For the purposes of this chapter, the Cartesian coordinates will be used interchangeably. 

% two polar coordinates $\mathbf{q}_{l/e} = \begin{bmatrix}r_{l/e} & \theta_{l/e}\end{bmatrix}^T$ and $\mathbf{q}_{r/e} = \begin{bmatrix} r_{r/e} & \theta_{r/e} \end{bmatrix}^T$,

%$ \dot{\mathbf{q}}_{l/e} = \begin{bmatrix} \dot{r}_{l/e} & \dot{\theta}_{l/e} \end{bmatrix}^T$ and $\dot{\mathbf{q}}_{r/e} = \begin{bmatrix} \dot{r}_{r/e} & \dot{\theta}_{r/e} \end{bmatrix}^T$ with equivalent Cartesian representations 

\begin{comment}
\begin{figure}[h!]
    \centering
    \includegraphics[width=0.75\linewidth]{figures/gap_diagram.png}
    \caption{Diagram of an example gap. Each gap is characterized by left and right gap point states comprised of positions $\mathbf{p}_{l/e}$ and $\mathbf{p}_{r/e}$ and velocities $\mathbf{v}_{l/e}$ and $\mathbf{v}_{r/e}$.}
    \label{fig:gap_diagram}
\end{figure}
\end{comment}

\begin{figure}[h!]
    \subfloat[Each gap $G$ is characterized by left and right gap point states comprised of positions $\mathbf{p}_{l/e}$ and $\mathbf{p}_{r/e}$ and velocities $\mathbf{v}_{l/e}$ and $\mathbf{v}_{r/e}$.]{ \includegraphics[width=0.45\textwidth]{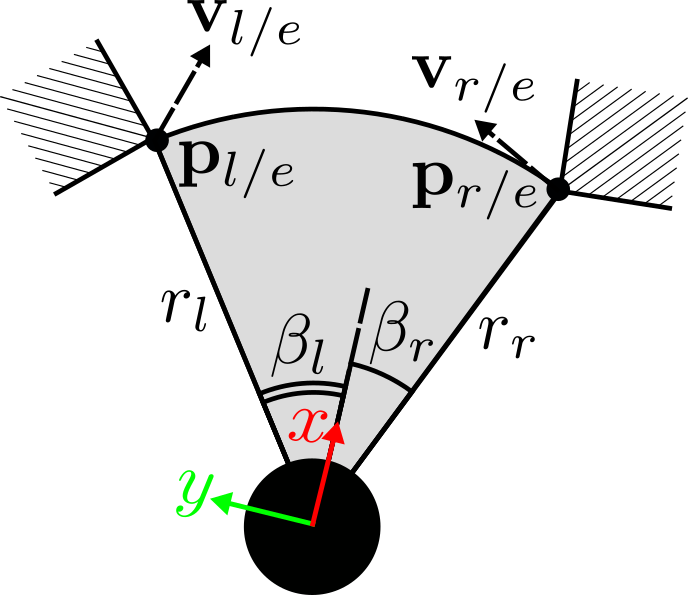}\label{fig:gap_diagram}}
    \quad
    \subfloat[Each ungap $U$ is built from the left gap point of a gap $G_i$, denoted as $\mathbf{p}^i_{l/e}$, and the right gap point of the next gap $G_{i+1}$, denoted as $\mathbf{p}^{i+1}_{r/e}$.]{\includegraphics[width=0.49\textwidth]{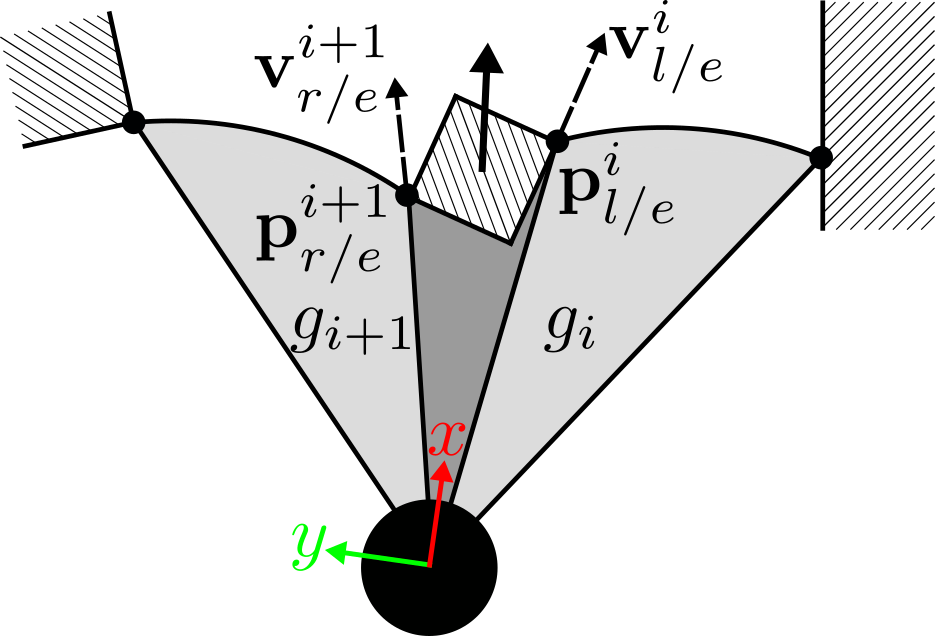}\label{fig:ungap_diagram}}
    \caption{Diagrams of (a) an example gap and (b) an example ungap.}
\end{figure}
\subsection{Ungap}
\textit{Definition (Ungap):} A polar region of obstacle space in $\mathbb{R}^2$ characterized by the area swept out by adjacent gap points $\mathbf{p}^{i+1}_{r/e}$ and $\mathbf{p}^{i}_{l/e}$.\\
%\[
%\mathbf{p}^{i+1}_{r/e} := (x^{i+1}_{r/e}, y^{i+1}_{r/e})
%\]
%and
%\[
%\mathbf{p}^{i}_{l/e} := (x^{i}_{l/e}, y^{i}_{l/e}).
%\] 
\begin{comment}
\begin{figure}[h!]
    \centering
    \includegraphics[width=0.99\linewidth]{figures/ungap_diagram.png}
    \caption{Diagram of an example ungap. Each ungap is built from the left gap point of a gap $g_i$, denoted as $\mathbf{p}^i_{l/e}$, and the right gap point of the next gap $g_{i+1}$, denoted as $\mathbf{p}^{i+1}_{r/e}$.}
    \label{fig:ungap_diagram}
\end{figure}
\end{comment}
Note that the two points that define an ungap come from two different adjacent gaps. The set of ungaps considered during planning is restricted to the ungaps that are (a) dynamic and (b) receding. The criteria used to determine these classifications are further detailed in Section \ref{sec:ungap_detection}.

\subsection{Gap Tube}
\textit{Definition (Gap Tube):} A sequence of gaps and gap lifespans that characterize how a gap evolves from time $t=0$ to the end of the local planning horizon $t=T$.\\
In this chapter, a gap tube will be expressed as 
\[
    \mathcal{T} = \{ (G_0, T_0), (G_1, T_1), \dots, (G_N-1, T_N-1) \}
\]

where $N$ is the number of unique gap models needed to characterize the gap tube during the local planning horizon and $T_0+T_1+\dots+T_{N-1}=T$. The process through which these gap tubes are constructed during planning is detailed in Section \ref{sec:gap_propagation}.
\begin{figure}[h!]
    \centering
    \includegraphics[width=0.99\linewidth]{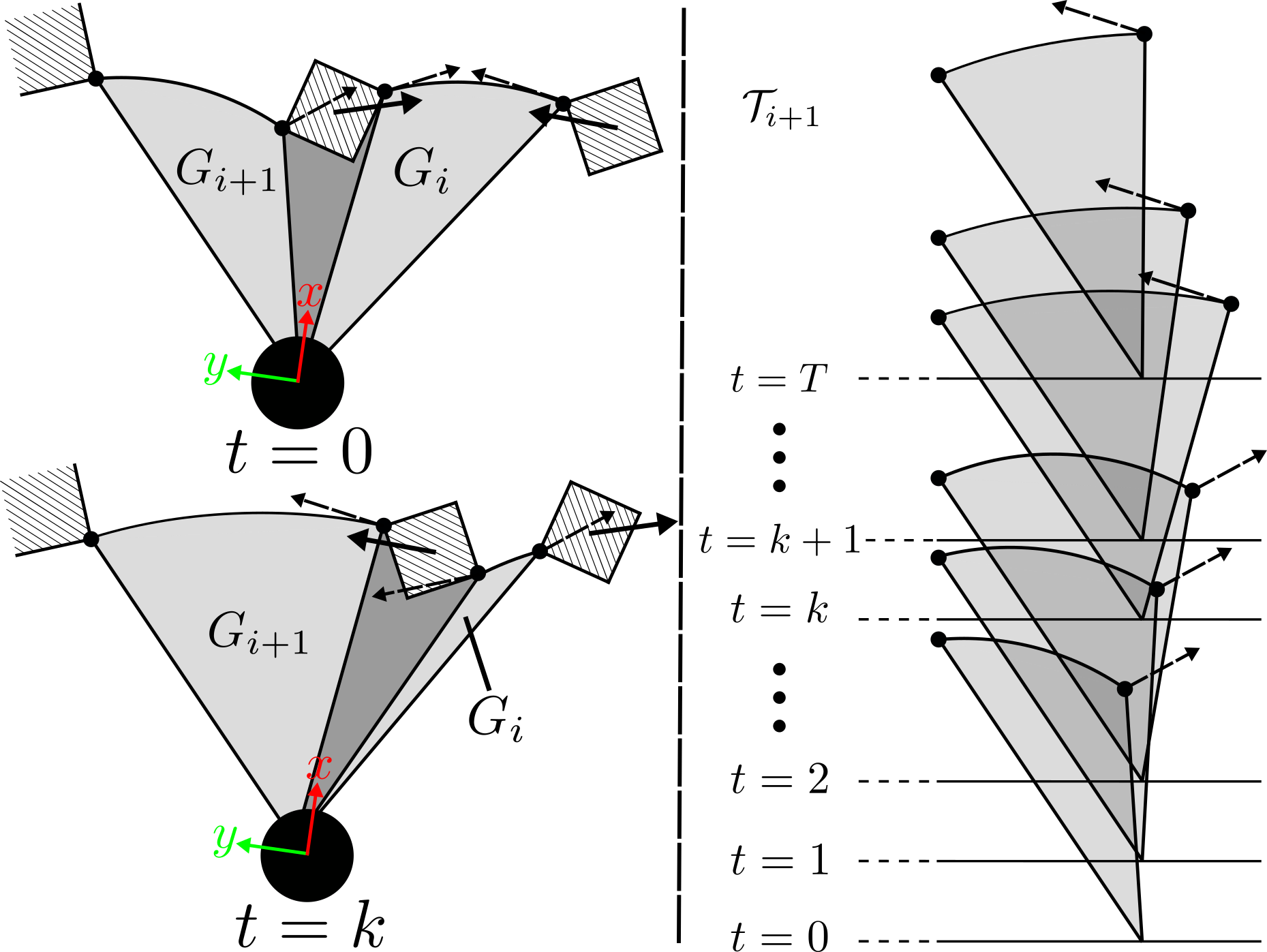}
    \caption{Diagram of an example gap tube. At $t=0$, the gap tube $\mathcal{T}_{i+1}$ begins with the gap $G_{i+1}$. However, at $t=k$, due to the two obstacles overlapping one another, the dynamics of the right point of gap $G_{i+1}$ change. In order to reflect this change in the dynamics, the gap tube $\mathcal{T}_{i+1}$ visualized on the right side of the figure captures this second gap with altered dynamics.}
    \label{fig:gap_tube_diagram}
\end{figure}

\section{Dynamic Gap Local Planning Module}
\label{sec:dynamic_gap}
\begin{comment}
    As input to the planner, access to a full FOV laser scan $\mathcal{L}$ comprised of a set of $n$ measurements of bearing and range tuples
\[
    \mathcal{L} = \{ (\beta_0, r_0), ... (\beta_{N-1}, r_{N-1}) \}
\]
is assumed. In practice, $\mathcal{L}$ starts at $\beta_0=-\pi$ and ends at $\beta_{N-1}=\pi$.
\end{comment}
\subsection{Gap Generation}
\label{sec:gap_generation}
The step of gap generation is an egocentric form of data clustering and environment abstraction that allows GBPs to condense a high-dimensional perceptual input such as a laser scan into a small set of free space regions that can be passed through during planning.

\subsubsection{Gap Detection}
\label{sec:gap_detection}
Gap detection involves the parsing of this scan $\mathcal{L}$ to obtain a set of detected gaps $\mathcal{G}^{\rm raw}_t$ that describe the free space of the local environment at the current time $t$. Gaps are classified as radial if they are formed from a significant, instantaneous change in range across consecutive scan points or as swept if they comprise a large interval of scan points with maximum detectable ranges.
%  Further details regarding how this gap detection policy is defined can be found at \cite{xu_potential_2021} or in Chapter \ref{ch:egonav}.

\subsubsection{Gap Simplification}
\label{sec:gap_simplification}
 Once this set of gaps has been extracted from the laser scan, an additional pass through $\mathcal{G}^{\rm raw}$ is performed to remove any redundant gaps and potentially merge adjacent gaps, yielding a set of simplified gaps $\mathcal{G}^{\rm simp}$ that represent the most s
 alient portions of free space in the local environment.
\begin{figure}[h!]
    \subfloat[Example set of raw and simplified gaps obtained from the gap detection policy in Sections \ref{sec:gap_detection} and \ref{sec:gap_simplification}. Blue points are the current laser scan, red arcs and points are the set of immediately detected raw gaps, and purple arcs and points are the simplified gaps.]{ \includegraphics[width=0.49\linewidth]{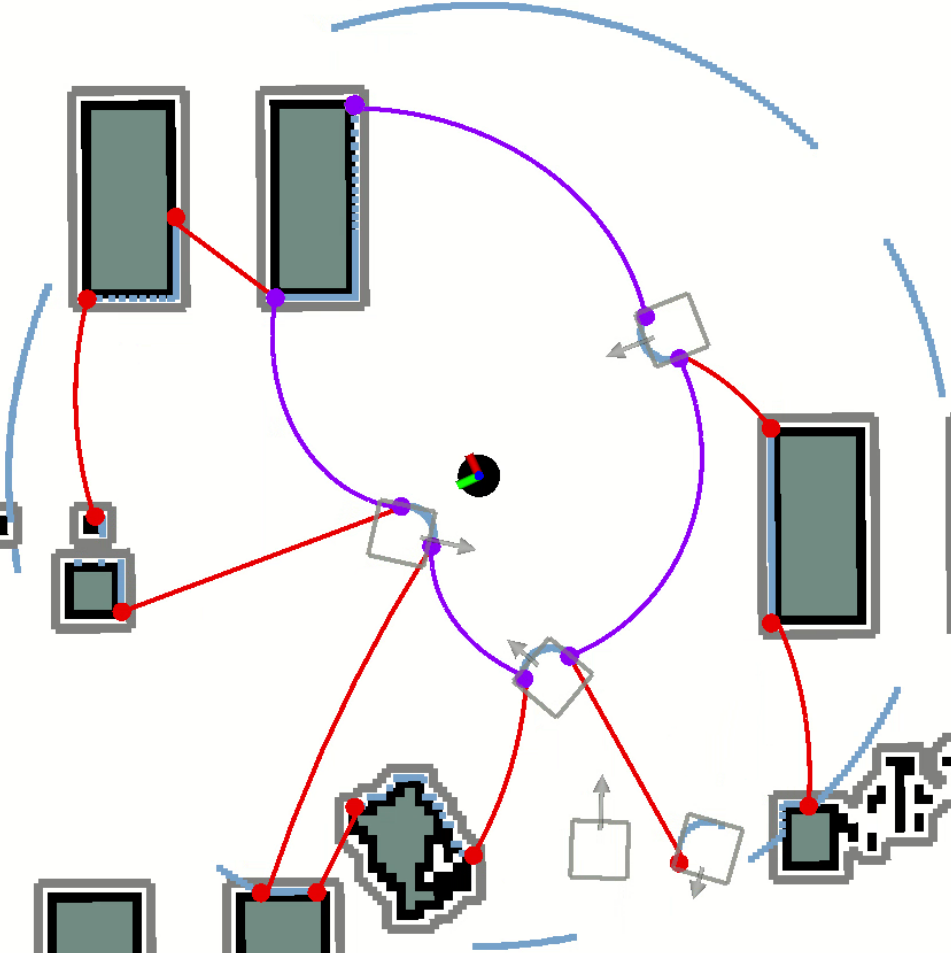}\label{fig:gap_simp_ex}}
    \quad
    \subfloat[Example set of raw gaps associated across timesteps as detailed in Section \ref{sec:gap_association}. As agent A passes in front of agent B at time $t$, the gaps attached to agent B are no longer visible and subsequently lost during gap association.]{\includegraphics[width=0.49\linewidth]{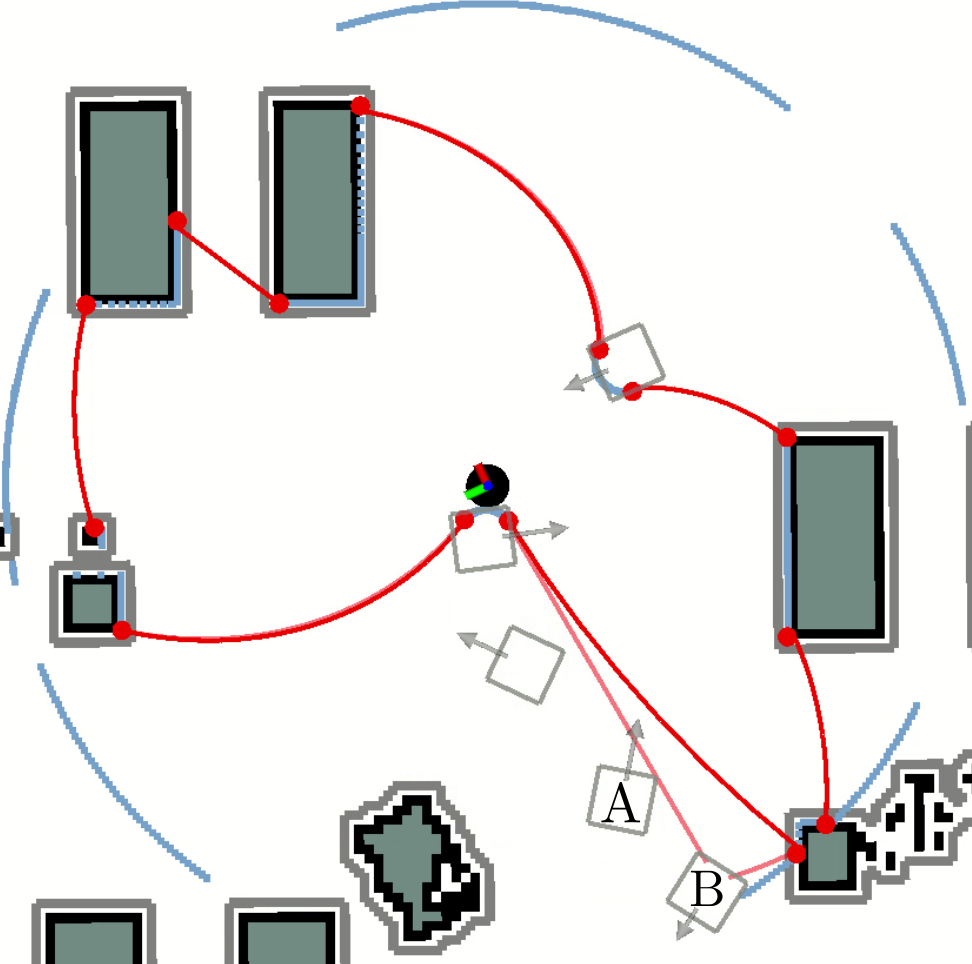}\label{fig:gap_assoc_ex}}
    \caption{Visualizations of (a) gap detection and simplification and (b) gap association.}
\end{figure}
% Blue points are the current laser scan, bolded red arcs are the set of raw gaps at time $t$. Semi-transparent red arcs are the set of gaps at time $t-1$.
\subsubsection{Ungap Detection}
\label{sec:ungap_detection}
A set of ungaps $\mathcal{U}^{\rm simp}$ can also be extracted from $\mathcal{G}^{\rm simp}$ to use during planning. To do so, the set of simplified gap points
\[
P^{\rm simp} = \{ \mathbf{p}_{r/e}^0, \mathbf{p}_{l/e}^0, ..., \mathbf{p}_{r/e}^N, \mathbf{p}_{l/e}^N \}
\]
is sorted in counterclockwise order by bearing. Then a series of checks are performed on adjacent pairs of points to determine if the pair $(\mathbf{p}_i, \mathbf{p}_j)$ forms an ungap. First, a velocity check is made to determine if the pair of points are dynamic and moving in roughly the same direction: \begin{equation} \label{eq:dist_check}
     \| \mathbf{v}_i \|_2 \geq v_{\rm min} \wedge \| \mathbf{v}_j \|_2 \geq v_{\rm min}
\end{equation}
and
\begin{equation} \label{eq:speed_check}
   \langle \mathbf{v}_i, \mathbf{v}_j \rangle > 0,
\end{equation}
where $v_{\rm min}$ is a minimum speed threshold to consider an obstacle as dynamic. If the two points meet the criteria of Equations \ref{eq:dist_check} and \ref{eq:speed_check}, then the points are attached to an ungap. Furthermore, if this ungap is determined to be receding, meaning that 
\begin{equation} \label{eq:receding_check}
     \langle \mathbf{p}_i, \mathbf{v}_i \rangle > 0 \wedge \langle \mathbf{p}_j, \mathbf{v}_j \rangle > 0,
\end{equation}
then this ungap is passed on to planning.
% and
%a distance check is performed to see if the two points could form an agent roughly the size of the ego-robot:
%\begin{equation}
%    \| \mathbf{p}_i - \mathbf{p}_j \|_2 \leq 4 r_{\rm infl}
%\end{equation}
% While this does assume a rough size limitation on the set of dynamic agents in the environment, the context of indoor navigation allows for such an assumption to be made. After this, an additional
\subsection{Gap Association}
\label{sec:gap_association}
The set of simplified gaps $\mathcal{G}^{\rm simp}$ captures the immediate free space, but in dynamic settings it is crucial to understand how this free space will evolve across the local planning horizon. Therefore, additional steps are taken to track gap points over time.

Association is performed on $P_{\rm simp}$ and represented as a RA problem where the cost is equivalent to the distance between points across consecutive steps, $P^{\rm simp}_{t-1}$ and $P^{\rm simp}_{t}$. This assignment problem is then solved with the HA \cite{kuhn_hungarian_1955}, producing a minimum distance mapping between $P^{\rm simp}_{t-1}$ and $P^{\rm simp}_{ t}$. If the distance between two associated points exceeds a threshold $\tau_{\rm assoc}\in\mathbb{R}^+$, then that point-to-point association is discarded and a new prediction model is instantiated for the respective point at time $t$.
%At this stage, each gap $G \in \mathcal{G}^{\rm simp}$ is characterized by a left gap point $\mathbf{p}_l = [x_l, y_l]^T$ and a right gap point $\mathbf{p}_r = [x_r, y_r]^T$. A
%\[
%P^{\rm simp} = \{ \mathbf{p}_r^0, \mathbf{p}_l^0, ..., %\mathbf{p}_r^N, \mathbf{p}_l^N \}.
%\]
\subsection{Gap Estimation}
\label{sec:gap_estimation}
The point-to-point associations provide an insight into how the set of gap points changes over time. This evolution is characterized by a second-order dynamics model with respect to the rotating ego-robot frame. 

Given the desire to perform gap estimation in the perception space, the prediction models must represent the gap points with respect to the local robot frame which is constantly changing. Therefore, the constant velocity dynamics model taken with respect to an inertial frame is augmented to allow for translations and rotations of the robot. The state vector is defined as
\begin{equation}
    \textbf{X}_s = \begin{bmatrix}
                \mathbf{p}_{s/e} \\
                \mathbf{v}_{s/e} \\
                \end{bmatrix} = 
                \begin{bmatrix}
                \mathbf{p}_s - \mathbf{p}_e \\
                \mathbf{v}_s - \mathbf{v}_e
                \end{bmatrix},	
\end{equation}
where $\mathbf{p}_{s/e} \in \mathbb{R}^2$ and $\mathbf{v}_{s/e} \in \mathbb{R}^2$ represent the position and velocity of the gap side $s$ (left or right) relative to the ego-robot $e$, respectively. The system dynamics are then
\begin{equation} \label{eq:rot_frame_dynamics}
    \dot{\textbf{X}}_s = \begin{bmatrix}
                \dot{\mathbf{p}}_{s/e} \\
                \dot{\mathbf{v}}_{s/e} \\
                \end{bmatrix} = 
                \begin{bmatrix}
                \mathbf{v}_{s/e} - \omega_e \times \mathbf{p}_{s/e}  \\
                \mathbf{a}_{s/e} - \omega_e \times \mathbf{v}_{s/e} \\
                \end{bmatrix},
\end{equation}
where $\omega_e$ represents the angular velocity of the ego-robot and $\mathbf{a}_{s/e}$ represents the linear acceleration of the gap side relative to the ego-robot. Gap points are assumed to travel at a constant velocity, which then simplifies Equation \ref{eq:rot_frame_dynamics} to
\begin{equation} \label{eq:rot_frame_dynamics_simp}
    \dot{\textbf{X}}_s = \begin{bmatrix}
                \dot{\mathbf{p}}_{s/e} \\
                \dot{\mathbf{v}}_{s/e} \\
                \end{bmatrix} = 
                \begin{bmatrix}
                \mathbf{v}_{s/e} - \omega_e \times \mathbf{p}_{s/e}  \\
                -\mathbf{a}_e - \omega_e \times \mathbf{v}_{s/e} \\
                \end{bmatrix}.
\end{equation}

This model is estimated with an extended KF given the nonlinear cross-product in Equation \ref{eq:rot_frame_dynamics_simp}. 

\begin{comment}
\begin{figure}[h!]
    \centering
    \includegraphics[width=0.99\linewidth]{figures/gap_assoc_and_est.png}
    \caption{Example association of raw gaps across timesteps. Color one is raw gaps at time t-1. Color two is scan at time t-1. Color three is raw gaps at time t. Color four is scan at time t. Numbers indicate gap point pairings at association.}
    \label{fig:gap_simp_ex}
\end{figure}
\end{comment}

\subsection{Gap Manipulation}
\label{sec:gap_manipulation}
Before the current set of gaps is passed on to trajectory generation, two pre-processing steps are taken in order to improve gap visibility and safety.

\subsubsection{Radial Gap Conversion}
\label{sec:radial_gap_conversion}
After the gap simplification policy defined in Section \ref{sec:gap_simplification}, it is still possible to have radial gaps in the current set of gaps. Radial gaps can cause safety risks due to their low visibility beyond the gap arc. A real-world example can be a hallway corner where it is difficult to determine if there is an upcoming obstacle beyond the corner. To mitigate this source of danger, remaining radial gaps are converted into swept gaps by pivoting the gap representation about the closer of the two gap points. Only gap points that are not attached to ungaps are converted so as to not neglect the dynamics of the environment, and the prediction models of all gap points that are altered through this radial gap conversion step are modified to reflect the new gap state.

\subsubsection{Gap Point Inflation}
\label{sec:gap_point_inflation}
\begin{SCfigure}
    \includegraphics[width=0.75\linewidth]{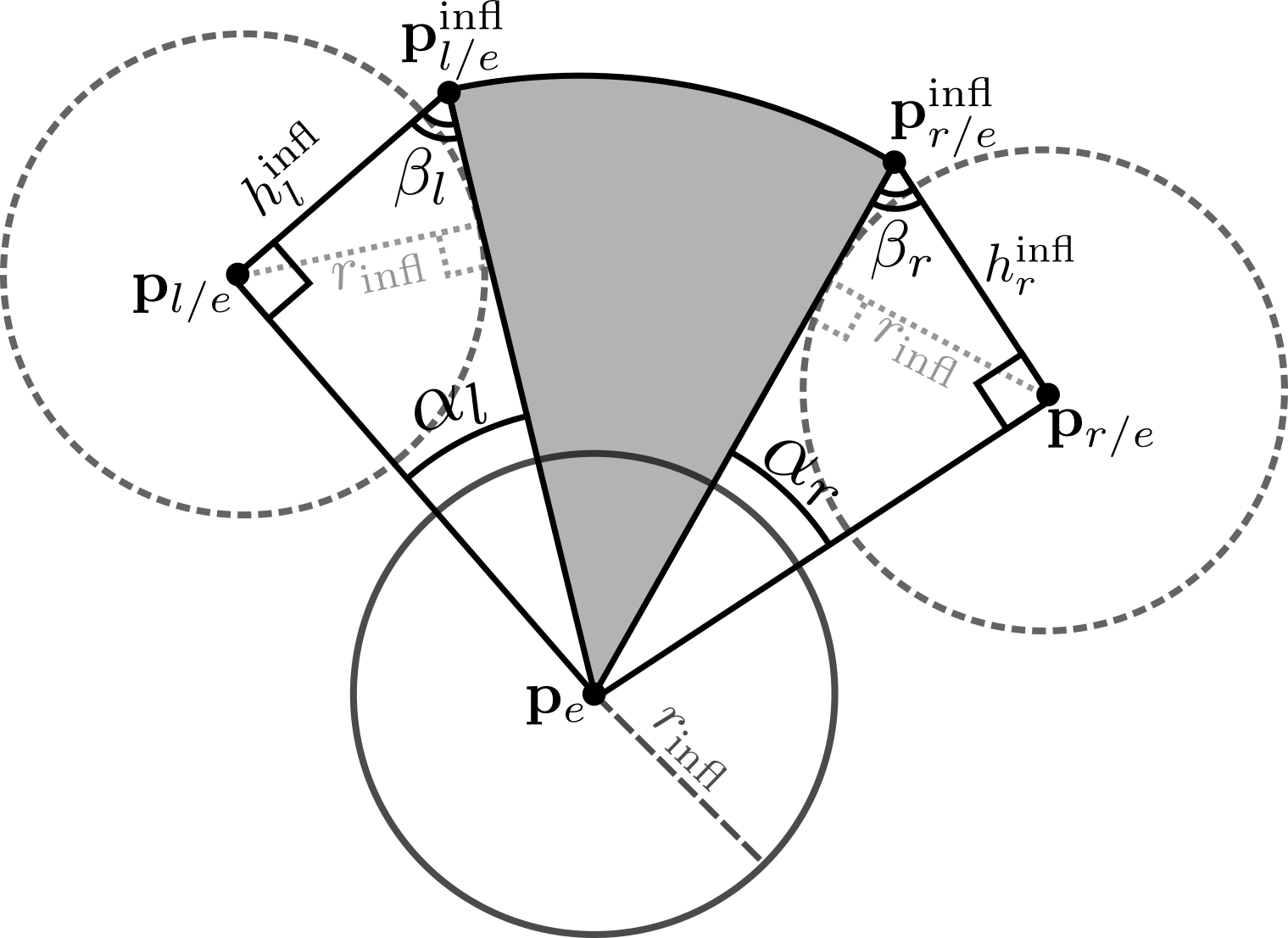}
    \caption{Diagram representing an example inflated gap. Here, the ego-robot point $\mathbf{p}_e$, the left gap point $\mathbf{p}_{l/e}$, and right gap point $\mathbf{p}_{r/e}$ are displayed along with their inflated counterpart $\mathbf{p}_{l/e}^{\rm infl},\mathbf{p}_{r/e}^{\rm infl}$. The inflation policy is identical for the both gap sides, with solely the rotation direction for the inflation point being the difference.}
    \label{fig:inflation_diagram}
\end{SCfigure}

Any real world robot that this planner is to be deployed on will have some finite size. Therefore, gap representations must be inflated inwards to compensate for the inscribed radius of the ego-robot. A diagram for how this inflation is performed is shown in Figure \ref{fig:inflation_diagram}. A user-defined inflation ratio $\tau_{\rm infl} \in [1, \infty)$ can be adjusted to modulate how conservative this step should be when inflating the gaps inward. First, the inflated inscribed robot radius is calculated as
\begin{equation}
    r_{\rm infl} = \tau_{\rm infl} \cdot r_{\rm inscr}.
\end{equation}
Then, the angle $\alpha$ that is formed between the line from $\mathbf{p}_e$ to the gap side point $\mathbf{p}_s$ and the line originating from $\mathbf{p}_e$ that is tangent to a circle of radius $r_{\rm infl}$ with its center at $\mathbf{p}_s$ is calculated as
\begin{equation}
    \alpha_s = \arcsin{(\frac{r_{\rm infl}}{\| \mathbf{p}_s \|})}.
\end{equation}
The remaining angle $\beta$ in the triangle formed by $\mathbf{p}_e$, $\mathbf{p}_s$, and the inflated gap side point $\mathbf{p}_s^{\rm infl}$ can be calculated as
\begin{equation}
    \beta_s = \frac{\pi}{2} - \alpha_s.
\end{equation}
Finally, the distance $h_{\rm infl}$ from the original gap side point $\mathbf{p}_s$ and the inflated gap side point $\mathbf{p}_s^{\rm infl}$ can be calculated as
\begin{equation}
    h^{\rm infl}_s = \frac{r_{\rm infl}}{\sin(\beta_s)}.
\end{equation}
\begin{comment}
\begin{figure}[h!]
    \centering
    \includegraphics[width=0.65\linewidth]{figures/inflation_diagram.png}
    \caption{Diagram representing an example inflated gap. Here, the ego-robot point $\mathbf{p}_e$, the left gap point $\mathbf{p}_{l/e}$, and right gap point $\mathbf{p}_{r/e}$ are displayed along with their inflated counterpart $\mathbf{p}_{l/e}^{\rm infl},\mathbf{p}_{r/e}^{\rm infl}$. The inflation policy is identical for the both gap sides, with solely the rotation direction for the inflation point being the difference.}
    \label{fig:inflation_diagram}
\end{figure}
\end{comment}
This manipulation step generates a set of manipulated gaps $\mathcal{G}^{\rm manip}$ which can be evaluated for feasibility.

\subsection{Gap Feasibility Analysis}
\label{sec:gap_feasibility_analysis}
In static settings, gap feasibility analysis is zeroth-order calculation. Determining whether or not a gap can be planned through is purely a geometric consideration: can the ego-robot fit within the gap? However, the transition to dynamic environments makes gap feasibility analysis a higher-order calculation.

In static settings, the currently available free space will remain as free space (and the same applies for obstacle space) for all time. In dynamic settings, free space can turn into obstacle space, and vice-versa. A gap currently exists, but might not in the future. A gap may not exist now, but will be created in the future. To account for these scenarios, the presently existent gaps are propagated forward in time to understand how they evolve, potentially changing shape or dynamics. By pruning away infeasible gaps, this step conserves the energy of the robot and avoids potentially dangerous gaps through which it would be difficult, if not impossible, for the robot to pass. 
% , but it also 
% not only 
%  of free space 
% for which planning will be performed
%\[
%P^{\rm manip} = \{ \mathbf{p}_l^0, \mathbf{p}_r^0, ..., \mathbf{p}_l^N, \mathbf{p}_r^N \}.
%\]
\subsubsection{Gap Propagation}
\label{sec:gap_propagation}
To understand the behavior of gaps over the local planning horizon, gap models (Eq. \ref{eq:rot_frame_dynamics_simp}) are integrated forward under the constant velocity assumption. In order to remove the ego-robot motion from the gap state and solely analyze the motion of the gap itself, the \textit{gap-only} dynamics are recovered from the prediction models by adding the ego-robot’s velocity $\mathbf{v}_e$ to the relative velocity estimate $\mathbf{v}_{s/e}$ to obtain the gap-only velocity $\mathbf{v}_s$. This gap-only velocity is then used during propagation. The general workflow is detailed below and given in Algorithms $\ref{alg:gap_point_propagation}$ and $\ref{alg:extract_propagated_gaps}$.

\textbf{Gap Point Propagation:} First, the set of points from the manipulated gaps $P^{\rm manip}$ is collected and propagated forward (Algorithm \ref{alg:gap_point_propagation}, Line \ref{line:propagate}). The propagated points are sorted in counterclockwise order by bearing (Line \ref{line:sort}), and these sorted points are passed on to Algorithm \ref{alg:extract_propagated_gaps} to extract a new set of gaps at that future timestep. The extracted gaps from consecutive timesteps $k-1$ and $k$ are associated (Line \ref{line:associate}) through a Rectangular Assignment problem that is slightly modified from the one in Section \ref{sec:gap_association} in order to assign the propagated gaps from $t_{k-1}$ to the gaps at $t_k$. Lastly, these newly associated gaps are merged into a temporal \textit{gap tube} that captures how a gap at time $t=0$ evolves in the local environment to time $t=T$ 
(Line \ref{line:merge}). This set of gap tubes $\mathcal{T}^{\rm prop}$ is evaluated for feasibility as detailed in Section \ref{sec:pursuit_guidance_analysis}.
\begin{algorithm}
	\caption{Gap Point Propagation}  \label{alg:gap_point_propagation}
	\begin{algorithmic}[1]
            \State \textbf{Given:} Manipulated gap points $P^{\rm manip}$ 
            \State \textbf{Return:} Set of propagated gap tubes $\mathcal{T}^{\rm prop}$
            \State \textit{/* Initialization */}
            \State $\mathcal{T}^{\rm prop} = \{ \}$
            \State $\mathcal{G}^{\rm prop}_{k-1} = \mathcal{G}^{\rm manip}$
            \State \textit{/* Propagation */}
            \For {each timestep $t_k$ in planning horizon $T$}
                \State Propagate $P^{\rm manip}$ to time $t_k$ \label{line:propagate}
                \State Sort $P^{\rm manip}$ in counterclockwise order by bearing \label{line:sort}
                \State Extract propagated gaps $\mathcal{G}^{\rm prop}_{k}$ from sorted $P^{\rm manip}$ \Comment Algorithm \ref{alg:extract_propagated_gaps}
                \State Associate propagated gaps $\mathcal{G}^{\rm prop}_k$ with $\mathcal{G}^{\rm prop}_{k-1}$ \label{line:associate}
                \State Merge associated propagated gaps $\mathcal{G}^{\rm prop}_k$ into $\mathcal{T}^{\rm prop}$ \label{line:merge}
                % \Comment Algorithm \ref{alg:associate_propagated_gaps}
                \State $\mathcal{G}^{\rm prop}_{k-1}$ = $\mathcal{G}^{\rm prop}_k$
            \EndFor
	\end{algorithmic} 
\end{algorithm}

\begin{figure}[h!]
    \subfloat[Gap propagation for two crossing agents. The bold colored arcs labeled A and B are the two gaps detected at time $t=0$. The lines of black points are the gap points propagated outwards from $t=0$ to $t=T$. The transparent colored arcs represent the gaps formed at future time steps. Available gaps are represented by solid arcs and unavailable gaps are represented by dashed arcs. At time $t=k-2$, the cyan arc is predicted to close, signaling that this gap is no longer available. At time $t=k$, the cyan gap points change dynamics, but the gap remains unavailable, marked by a second dashed arc. At $t=k$, the magenta arc changes gap dynamics but remains available.]{ \includegraphics[width=0.49\linewidth]{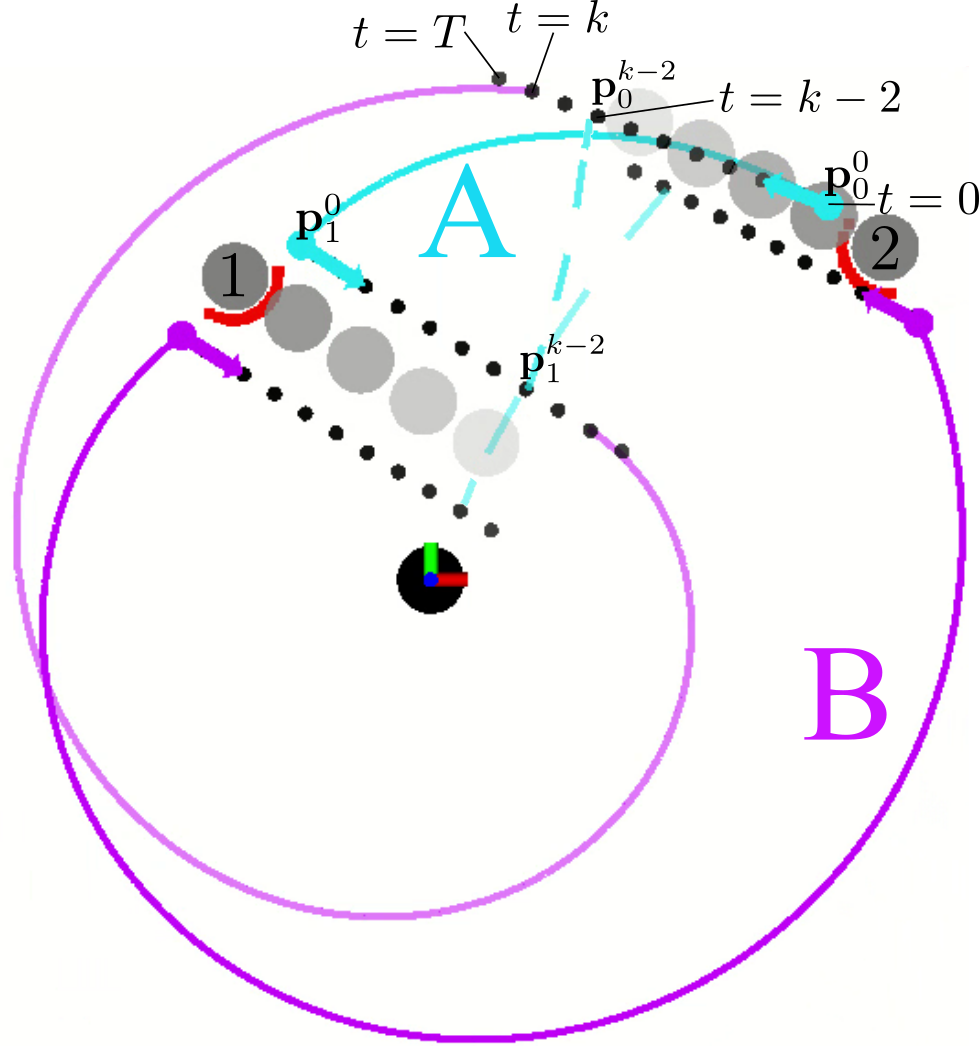}\label{fig:closing_and_reopening_example}}
    \quad
    \subfloat[Gap propagation for two translating agents (labeled $1$ and $2$) and a third interrupting agent (labeled $3$). The bold colored arcs labeled A -C are the three gaps detected at time $t=0$. At time $t=k-1$, the magenta arc is predicted to close, signaling that this gap is no longer available. At time $t=k$, the cyan gap is interrupted by agent $3$, marked by a transparent cyan arc, but the gap remains available at this time. The magenta arc changes gap dynamics but continues to be unavailable as well Lastly, the blue gap changes dynamics but continues to be available.]{\includegraphics[width=0.49\linewidth]{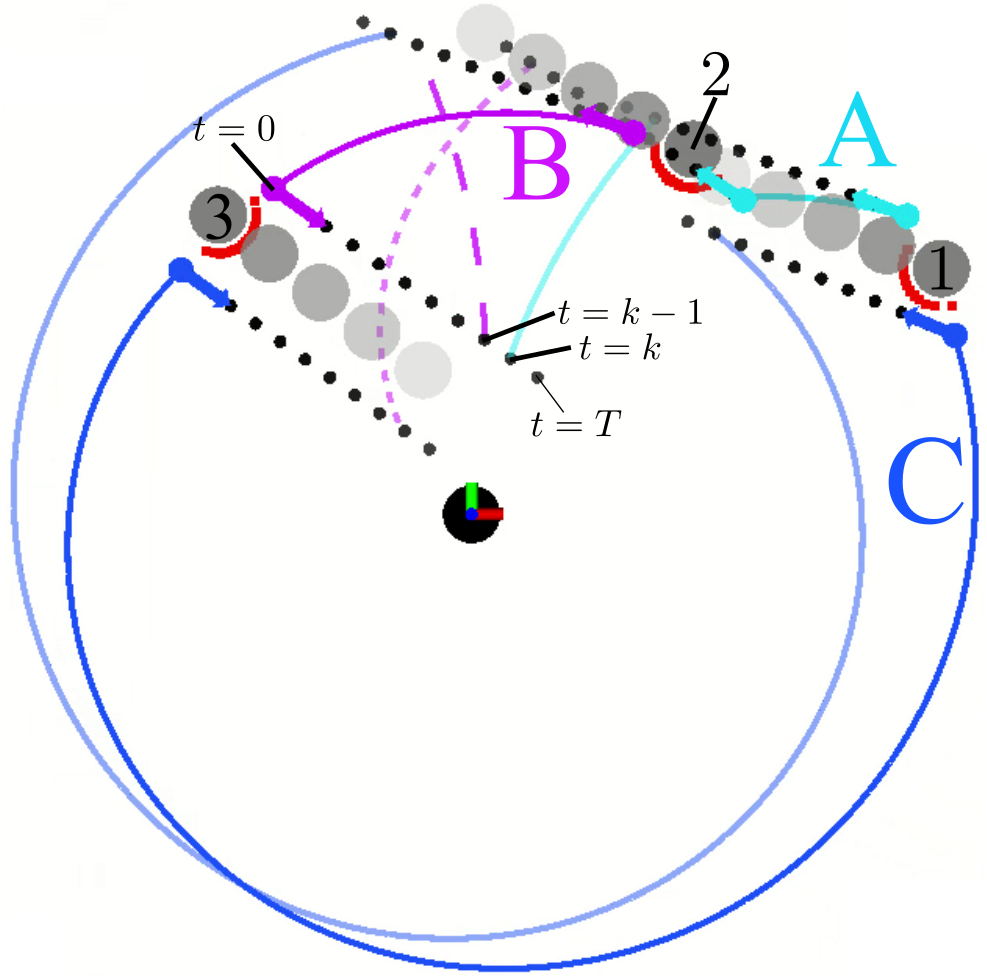}\label{fig:interrupting_diagram}}
    \caption{Visualizations of gap propagation algorithm.}
\end{figure}

\textbf{Propagated Gap Extraction:} Algorithm \ref{alg:extract_propagated_gaps} shows how a set of gaps at future time $t_k$ are extracted from a set of propagated gap points. This algorithm iterates through the ordered gap points (Lines $\ref{line:iter1} - \ref{line:iterN}$) 
in search of consecutive points for which $\mathbf{p}^{\rm manip}_i$ represents a right gap point and $\mathbf{p}^{\rm manip}_j$ represents a left gap point. One example of this would be the pair of points $\mathbf{p}^0_0$ and $\mathbf{p}^0_1$ in Figure \ref{fig:closing_and_reopening_example}. If the opposite occurs, and $\mathbf{p}^{\rm manip}_i$ represents a left gap point while $\mathbf{p}^{\rm manip}_j$ represents a right gap point, this represents a gap that exists in the environment but is currently unavailable due to obstacles passing across one another. An example of an unavailable gap is the pair of points $\mathbf{p}^{k-2}_0$ and $\mathbf{p}^{k-2}_1$ in Figure \ref{fig:closing_and_reopening_example}. This unavailable gap is formed when agents $1$ and $2$ that initially form gap A pass across each other at the timestep $t_{k-2}$. A new gap will appear at a future timestep, so it is important to store this information for gap propagation. Ungap IDs are leveraged in this step to ensure that neighboring gap points that form an ungap are not mistaken for an unavailable gap. 

\begin{algorithm} 
	\caption{Propagated Gaps Extraction} \label{alg:extract_propagated_gaps}
	\begin{algorithmic}[1]
            \State \textbf{Given:} Manipulated gap points $P^{\rm manip}$ 
            \State \textbf{Return:} Set of propagated gaps $\mathcal{G}^{\rm prop}_{k}$
            \State \textit{/* Initialization */}
            \State $N$ = $| P^{\rm manip} |$
            \State $\mathcal{G}^{\rm prop}_{k} = \{ \}$
            \State $i_0 = $ index of first right gap point
            \State $i \leftarrow 0$
            \State \textit{/* Extraction */}
            \For {$i = 0$ to $N$} $\label{line:iter1}$
                \State $\mathbf{p}^{\rm manip}_i$ = $P^{\rm manip}(i_0 + i)$
                \If{$\mathbf{p}^{\rm manip}_i$ is not assigned to a gap}
                    \State $l_i\leftarrow$ gap side for $\mathbf{p}^{\rm manip}_i$
                    \State $u_i\leftarrow$ ungap ID for $\mathbf{p}^{\rm manip}_i$
                    \For {$\Delta i = 1$ to $N$} $\label{line:iter2}$
                        \State j = ($i + \Delta i$) \% N
                        \State $\mathbf{p}^{\rm manip}_j$ = $P^{\rm manip}(i_0 + j)$
                        \If{$\mathbf{p}^{\rm manip}_j$ is not assigned to a gap}
                            \State $l_j\leftarrow$ gap side for $\mathbf{p}^{\rm manip}_j$
                            \State $u_j\leftarrow$ ungap ID for $\mathbf{p}^{\rm manip}_j$
                            \If{$l_i \neq l_j$ and $u_i \neq u_j$} $\label{line:iterN}$
                                \If{$l_i$ is right} \Comment{Right to left: available gap}
                                    \State Add an available gap between $\mathbf{p}^{\rm manip}_i$ and $\mathbf{p}^{\rm manip}_j$ to $\mathcal{G}^{\rm prop}_{k}$
                                \Else \Comment{Left to right: unavailable gap}
                                    \State Add an unavailable gap between $\mathbf{p}^{\rm manip}_i$ and $\mathbf{p}^{\rm manip}_j$ to $\mathcal{G}^{\rm prop}_{k}$
                                \EndIf
                            \EndIf
                        \EndIf
                    \EndFor
                \EndIf
            \EndFor
	\end{algorithmic} 
\end{algorithm} 

Once a set of propagated gaps $\mathcal{G}^{\rm prop}_{k}$ has been extracted for future time $t_k$, a gap association step is performed between $\mathcal{G}^{\rm prop}_{k-1}$ and $\mathcal{G}^{\rm prop}_{k}$. This step is identical to the Rectangular Assignment problem discussed in Section \ref{sec:gap_association}, the only difference here being that entire gaps are associated instead of individual gap points. Since gaps are associated, the distance metric used is the sum of squared differences between the points of the two gaps $G_i$ and $G_j$:

\begin{equation}
    d_{ij} = \| \mathbf{p}_{l, i} - \mathbf{p}_{l, j} \|_2^2 + \| \mathbf{p}_{r, i} - \mathbf{p}_{r, j} \|_2^2.
\end{equation}

From this association step, there can be several outcomes. If the associated gaps $G_i$ and $G_j$ share the same left and right points, then this gap was not interrupted by any other gaps during propagation and it can be left alone. However, if two associated gaps have different points, this indicates some form of a gap interruption. This interruption might be caused by a gap closing and eventually re-opening (as shown in Figure \ref{fig:closing_and_reopening_example}), or by one gap being propagated into the space of a different gap (as shown in Figure \ref{fig:interrupting_diagram}). In either of these scenarios, the gap dynamics change, and a new gap must be instantiated within this particular gap tube.

% Available gaps are represented by solid arcs and unavailable gaps are represented by dashed arcs. 
% The transparent colored arcs represent the gaps formed at future time steps. 
% The lines of black points are the gap points propagated outwards from $t=0$ to $t=T$.

\subsubsection{Pursuit Guidance Analysis}
\label{sec:pursuit_guidance_analysis}
In this section, the guidance law-based trajectory generation scheme that dynamic gap employs in order to determine if a gap tube is kinematically feasible will be outlined. Guidance laws \cite{shneydor_missile_1998} comprise a set of kinematic equations and feedback control laws that define collision course behavior between a pursuer and a target. While commonly affiliated with older forms of missile guidance, these laws have also seen use in many robotics applications \cite{noauthor_implementation_nodate, wellhausen_artplanner_2023, fiorini_motion_1998}.

Among the more established guidance laws, the two geometrical rules of Pure Pursuit (PP) and Parallel Navigation (PN) are the most popular. The PP rule, sometimes referred to as pursuit guidance, has the pursuer direct their velocity vector towards the target at all times, always keeping the target within the pursuer’s line of sight. PP has seen a great deal of attention due to its simplicity \cite{bernhart_polygons_1959, bernhart_curves_1959, bruckstein_why_1993}, but this guidance law only leads to a collision if the pursuer is capable of traveling at a speed faster than that of the target. % If the pursuer can not travel as fast as the target, this guidance law will yield a tail-chase.

The PN, or constant bearing, rule \cite{rajasekhar_fuzzy_2000, ulybyshev_terminal_2005} has the pursuer direct their velocity vector such that the direction of the line of sight between the pursuer and target remains constant while the distance between them decreases. This geometrical rule is capable of yielding collision course conditions even if the pursuer is traveling slower than the target. Furthermore, for a non-maneuvering target, meaning a target that is not changing its speed nor its heading direction, PN is the optimal guidance law which yields a minimum intercept time. For a single gap tube, each gap in the gap tube is evaluated independently for feasibility. A gap tube is feasible if all gaps within the tube are feasible.

\begin{SCfigure}
    \includegraphics[width=0.50\linewidth]{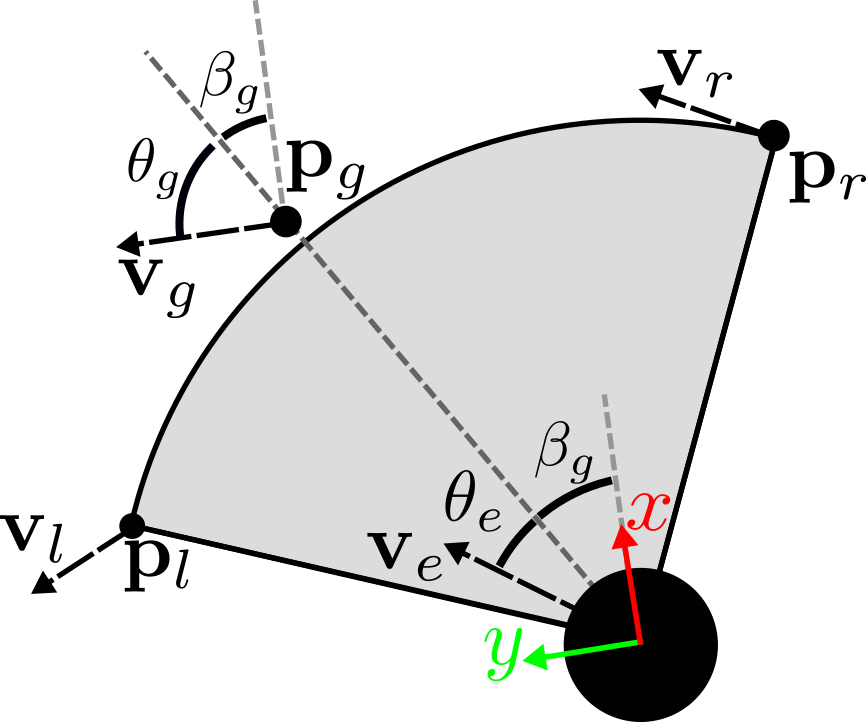}
    \caption{Diagram for guidance law notation for a single gap. The bearing $\beta_g$ denotes the bearing at which the gap goal is at for $t=0$. The angles $\gamma_g = \theta_g + \beta_g$ and $\gamma_e = \theta_e + \beta_g$ represent the directions in which the gap goal position and the robot position move for interception.}
    \label{fig:pursuit_guidance_diagram}
\end{SCfigure}

\begin{comment}
\begin{figure}[h!]
    \centering
    \includegraphics[width=0.65\linewidth]{figures/pursuit_guidance_diagram.png}
    \caption{Diagram for guidance law notation for a single gap. The bearing $\beta_g$ denotes the bearing at which the gap goal is at for $t=0$. The angles $\gamma_g := \theta_g + \beta_g$ and $\gamma_e := \theta_e + \beta_g$ represent the directions in which the gap goal position and the robot position move for interception.}
    \label{fig:pursuit_guidance_diagram}
\end{figure}
\end{comment}

This feasibility analysis employs the PN geometric rule \cite{shneydor_missile_1998} which assumes a constant gap goal speed $v_g$ as well as a constant ego-robot speed $v_e$. This policy then defines the bearing $\gamma_e = \theta_e + \beta_g$ towards which $v_e$ will be applied. With relation to Figure \ref{fig:pursuit_guidance_diagram}, the PN policy is defined as $\dot{\beta}_g = 0, \dot{r}_g < 0$, where
\begin{equation} \label{eq:par_nav_conditions}
    \begin{bmatrix}
                \dot{\beta}_g \\
                \dot{r}_g \\
                \end{bmatrix} = 
                \begin{bmatrix}
                \frac{v_g \sin(\theta_g) - v_e \sin(\theta_e)}{r_g} \\
                v_g \cos(\theta_g) - v_e \cos(\theta_e) \\
                \end{bmatrix},
\end{equation}
and $r_g := \| \mathbf{p}_g \|$. In order for $\dot{\beta}_g = 0$ to hold,
\begin{equation}
    v_g \sin(\theta_g) = v_e \sin(\theta_e).
\end{equation}
In order for $\dot{r}_g < 0$, it must be that
\begin{equation}
    v_e \cos(\theta_e) < v_g \cos(\theta_g).
\end{equation}
Therefore, for a constant speed ratio $K := v_e / v_g$, it follows that the gap goal position can be attained if
\begin{equation} \label{eq:theta_e}
    \sin(\theta_e) = \frac{\sin(\theta_g)}{K},
\end{equation}
and
\begin{equation}
    \cos(\theta_e) > \frac{\cos(\theta_g)}{K}.
\end{equation}
If these conditions can be met, then the ego-robot will intercept the goal position at the time
\begin{equation}
    t_{\rm intercept} = \frac{r_g^0}{v_g} \cdot \frac{1}{K \cdot \cos(\theta_e) - \cos(\theta_g)},
\end{equation}
where $r_g^0$ is equal to $r_g$ at $t=0$.
If these conditions can not be satisfied, then the given gap tube is deemed infeasible and discarded. If the conditions can be satisfied, but $t_f < t_{\rm intercept}$, meaning that the gap will cease to exist before the ego-robot can intercept the goal position, then the gap tube is also deemed infeasible and discarded. Gap tubes that satisfy this condition are added to $\mathcal{T}^{\rm feas}.$

\subsubsection{Proof of Collision-Free Passage} \label{sec:proof} 

%  constant velocity - not necessarily an assumption, just the policy we enact
% point-mass - we can manipulate gap

\textit{Theorem 1}: Given 
\begin{enumerate}
    \item A first-order holonomic ego-robot,
    \item A feasible manipulated gap with constant velocity left and right points $\mathbf{p}_l, \mathbf{p}_r$,
    \item An isolated local environment meaning that no other gaps will enter the gap in focus during the local time horizon,
\end{enumerate}
then performing the PN policy towards the gap goal point $\mathbf{p}_g$ will yield collision-free gap passage.\\
\textit{Proof:} Let the gap goal point $\mathbf{p}_g$ and velocity $\mathbf{v}_g$ be defined as a convex combination of the left and right gap point states,

\begin{equation}
\begin{split}
    \mathbf{p}_g & = \kappa \mathbf{p}_l + (1 - \kappa) \mathbf{p}_r, \\
    \mathbf{v}_g & = \kappa \mathbf{v}_l + (1 - \kappa) \mathbf{v}_r, \hspace{0.5cm} \kappa \in [0, 1].
\end{split}
\end{equation}
If the gap in focus possesses an angular span of greater than $\pi$ radians, meaning that the gap is nonconvex and a convex combination may lie outside of the gap, the gap span can be artificially reduced to two points within the original gap span that form a convex polar triangle. It follows that
\begin{equation}
    \beta_g = \arctan(\mathbf{p}_g) = \arctan(\kappa \mathbf{p}_l + (1 - \kappa) \mathbf{p}_r).
\end{equation}
Without loss of generality (by rotating the egocentric frame to align with the center of the initial gap), $\beta_g \in [\beta_r, \beta_l]$ given that $\arctan$ is a monotonically increasing function. Following the same argument for
\begin{equation}
    \gamma_g = \arctan(\mathbf{v}_g) = \arctan(\kappa \mathbf{v}_l + (1 - \kappa) \mathbf{v}_r),
\end{equation}
it can be seen that $\gamma_g \in [\gamma_r, \gamma_l]$. Given that
\begin{equation}
    \gamma = \beta + \theta,
\end{equation}
it follows that $\theta_g \in [\theta_r, \theta_l]$.
From Equation \ref{eq:theta_e},
\begin{equation}
    \theta_e = \arcsin{( \frac{\sin{\theta_g}}{K})},
\end{equation}
and given that $\arcsin$ is also a monotonically increasing function, this means that $\theta_{e} \in [ \theta_{e / r}, \theta_{e / l}]$ where $\theta_{e}, \theta_{e / l}, \theta_{e / r}$ are the bearings at which the ego-robot must direct its velocity at to intercept the gap goal, the left gap point, and the right gap point, respectively. This indicates that under the PN policy, the ego-robot will intercept the gap goal point between the left and right gap points, therefore performing collision-free gap passage. $\qed$

While it is clear that in practice, few gaps will satisfy these constant velocity and isolated environment assumptions, the trajectory generation scheme for the proposed planner is still built upon a collision-free guarantee. Supplementary modules including gap propagation, scan propagation, and safety filters are all employed to mitigate risk in situations where assumptions are violated.

% \subsubsection{Inflation Policy}
% \label{sec:inflation_policy}

% \textcolor{red}{Gap inflation policy}

% \subsubsection{Radial Gap Conversion} 
% \label{sec:radial_gap_conversion}

\subsection{Gap Trajectory Generation}
\label{sec:gap_trajectory_generation}

% \subsubsection{Gap Goal Placement}
% \label{sec:gap_goal_placement}
\textbf{Gap Goal Placement:} For a given gap $G_i$ within a gap tube $\mathcal{T}_j$, the gap goal state $\mathbf{X}_g$ is a convex combination of the left and right gap point states. The position of a local waypoint $\mathbf{p}*$ from the global trajectory $\xi^{\rm global}$ can be used to bias this gap goal towards one side. For ungaps, the gap goal is inflated inwards to ensure that it can be reached without colliding with the obstacle creating the ungap.\\
% \subsubsection{Gap Trajectory Rollout}
% \label{sec:gap_trajectory_rollout}
\textbf{Gap Trajectory Rollout:} Gap tube trajectories are obtained by integrating the ego-robot position forward along the intercepting heading $\gamma_e$ resulting from the pursuit guidance analysis in Section \ref{sec:pursuit_guidance_analysis}. Ungap trajectories are synthesized in the same manner.
\subsection{Gap Trajectory Scoring}
\label{sec:gap_trajectory_scoring}
\subsubsection{Pose-wise Scoring}
\label{sec:pose_wise_scoring}
In order to determine which trajectory to track, each trajectory is evaluated by an egocentric pose-wise cost based on proximity to local obstacles and a terminal pose cost based on proximity to a local waypoint along the global plan in order to encourage progress toward the global goal. 

%  = \{\mathbf{p}_i^0, \mathbf{p}_i^1, ..., \mathbf{p}_i^N \} 

The cumulative cost for a trajectory $\xi$ is
\begin{equation} \label{eq:traj_cost}
\mathcal{J}(\xi) = w \| \mathbf{p}[N] - \mathbf{p}^* \| + \frac{1}{N}\sum_{k=1}^{N} C( d(\mathbf{p}[k]; \mathcal{L}[k])) \\
\end{equation}
where $d(\mathbf{p}; \mathcal{L})$ is the distance from the pose $\mathbf{p}$ to the laser scan $\mathcal{L}$, $w \in \mathbb{R}^+$ is a weighting factor, $\mathbf{p}^*$ is a local waypoint along the global path, and
\begin{equation}
    C(d) = \begin{cases}
                \infty, & d \leq r_{\rm infl} \\
			c_{\rm obs} e ^ {-w_2 (d - r_{\rm infl})}, & r_{\rm infl} < d < r_{\rm max} \\
                0, & r_{\rm max} \leq d
		  \end{cases},
\end{equation} 
where $c_{\rm obs}, w_2 \in \mathbb{R}^+$ are weighting factors, $r_{\rm infl}$ is the inflated radius of the ego-robot, and $r_{\rm max}$ is the maximum distance that is penalized. Pose-wise scores are averaged so as to not bias selection towards shorter trajectories.

\subsubsection{Dynamic Scan Propagation}
\label{sec:dynamic_scan_propagation}
\begin{SCfigure}
    \includegraphics[width=0.50\linewidth]{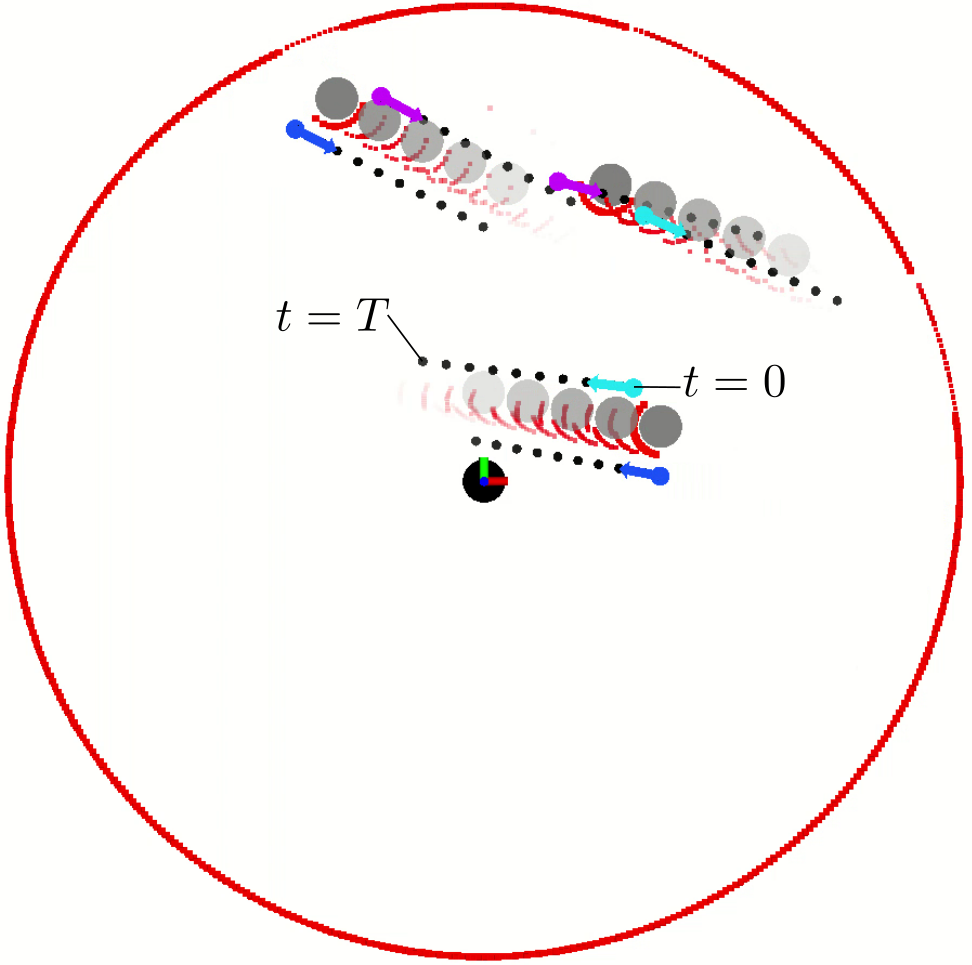}
    \caption{Visualization of dynamic scan propagation. The colored points are the gap points along with their estimated velocities. Similar gap points are used to propagate scan points from $t=0$ to $t=T$, represented by the increasingly transparent scan points.}
    \label{fig:dynamic_scan_propagation}
\end{SCfigure}

The trajectory cost formulation is adapted from \cite{xu_potential_2021}, with one key difference: pose-wise scoring requires a laser scan for each timestep $t$ along the trajectory. However, these future laser scans are not directly accessible. In practice, when gaps are propagated forward in time, a set of propagated laser scans are recovered and stored for later scoring by estimating the laser scan dynamics $\dot{\mathcal{L}}$. The algorithm used to back out these propagated scans from a set of predicted gaps is given in Algorithm \ref{alg:dynamic_scan_propagation}. An example set of propagated scans obtained during planning is visualized in Figure \ref{fig:dynamic_scan_propagation}.

\begin{algorithm}
	\caption{Dynamic Scan Propagation} \label{alg:dynamic_scan_propagation}
	\begin{algorithmic}[1]
            \State \textbf{Given:} Current set of raw gap points $P^{\rm raw}$
            \State \textbf{Given:} Current laser scan $\mathcal{L}_0$
            \State \textbf{Return:} Set of propagated laser scans $\{ \mathcal{L}_0, \mathcal{L}_1, \dots, \mathcal{L}_T  \}$
            \State \textit{/* Assignment */}
            \State $\dot{\mathcal{L}} = \{ \}$
            \For{$ (\beta_i, r_i) \in \mathcal{L}_0$ }
                \For{$\mathbf{p}_j \in P^{\rm raw}$}
                    \State $\beta_j = \arctan{(\mathbf{p}_j)}$
                    \If{$\beta_j \leq \beta_i$}
                        \State $\mathbf{p}_{\rm RHS} \leftarrow \mathbf{p}_j$ 
                    \Else % {$\beta_j \geq \beta_i$}
                        \State $\mathbf{p}_{\rm LHS} \leftarrow \mathbf{p}_j$ 
                        \State $\textbf{break}$
                    \EndIf
                \EndFor

                \If{areSimilar($\mathbf{p}_{\rm LHS}, \mathbf{p}_{\rm RHS}$)} \Comment{Equation \ref{eq:isSimilar}}
                    \State $\dot{\beta}_i \leftarrow (\dot{\beta}_{\rm LHS} + \dot{\beta}_{\rm RHS}) / 2 $ % Assign closer of two models to scan point
                    \State $\dot{r}_i \leftarrow (\dot{r}_{\rm LHS} + \dot{r}_{\rm RHS}) / 2 $
                \Else
                    \State $\dot{\beta}_i = 0$
                    \State $\dot{r}_i = 0$
                \EndIf
            \EndFor
            \State \textit{/* Propagation */}
            \For {each timestep $t_k$ in planning horizon $T$}
                \State $\mathcal{L}_k = \mathcal{L}_0 + \dot{\mathcal{L}} \cdot t_k$ \label{line:scan_prop}
            \EndFor
	\end{algorithmic} 
\end{algorithm} 

In short, a dynamics model for the laser scan $\dot{\mathcal{L}}$ is approximated by assigning gap point dynamics models to each scan point if the neighboring gap point models are sufficiently similar. In this algorithm, the condition areSimilar that determines if two gap point states $(\mathbf{p}_{\rm LHS}, \mathbf{v}_{\rm LHS})$ and $(\mathbf{p}_{\rm RHS}, \mathbf{v}_{\rm RHS})$ are similar is evaluated as

\begin{equation} \label{eq:isSimilar}
    \| \mathbf{v}_{\rm LHS} \|_2 \geq v_{\rm min} \wedge  \| \mathbf{v}_{\rm RHS} \|_2 \geq v_{\rm min} \wedge
    \mathbf{v}_{\rm LHS} \cdot \mathbf{v}_{\rm RHS} > 0.
\end{equation}

Given that propagated scan points can change both their bearing and range, care must be taken to ensure that propagated points are mapped to the correct scan point index.
% In line \ref{line:scan_prop} of Algorithm \ref{alg:dynamic_scan_propagation}, the notation for time derivatives is slightly abused. 

The highest scoring candidate trajectory is compared against the currently executing trajectory to determine if a trajectory change should occur. 

\begin{comment}
\begin{figure}[h!]
    \centering
    \includegraphics[width=0.65\linewidth]{figures/scan_propagation.png}
    \caption{Visualization of dynamic scan propagation. The colored points are the gap points along with their estimated velocities. Similar gap points are used to propagate scan points from $t=0$ to $t=T$, represented by the increasingly transparent scan points.}
    \label{fig:dynamic_scan_propagation}
\end{figure}
\end{comment}

\subsection{Gap Trajectory Comparison}
\label{sec:gap_trajectory_comparison}
The core idea behind safe hierarchical planners involves chaining together multiple safe local trajectories en route to the global goal. Therefore, a method of triggering a switch to a newly synthesized local trajectory must be defined. In this work, an event-based trajectory switching scheme is employed in which a trajectory switch only occurs if one of a set of conditions is met, as described below.

If the trajectory that is currently being tracked has either been completed (evaluated by proximity to the trajectory end or by the trajectory timing) or determined to be on a collision course, then a switch is triggered to the newly synthesized incoming trajectory. Similarly, if the gap that the currently tracked trajectory passes through has been determined to be infeasible as per Section $\ref{sec:gap_feasibility_analysis}$, a switch is triggered.

\subsection{Gap Trajectory Tracking}
\label{sec:gap_trajectory_tracking}
Once a local trajectory is chosen for tracking, a simple state feedback control law is deployed to track the trajectory.  

\subsection{Projection Operator}
\label{sec:projection_operator}
As a last resort safety filter to handle non-ideal circumstances including discrete time implementation and second-order dynamics, the proposed work also adapts the projection operator module which is detailed in \cite{xu_potential_2021}.

\begin{comment}
from \cite{xu_potential_2021} for which the potential function is defined as

\begin{equation}
    \psi(x) = \Bigl( \frac{r_{\rm min}}{d(x; \mathcal{L})} - \frac{r_{\rm min}}{r_{\rm nom}} \Bigr) \Big / \Bigl( 1 - \frac{r_{\rm min}}{r_{\rm nom}} \Bigr),
\end{equation}

where $r_{\rm min} < r_{\rm nom}$. 
\end{comment}

\section{Experimental Results}
\label{sec:experimental_results}

This planner is implemented as a C++ Robot Operating System (ROS) node through the \texttt{move\_base} package \cite{noauthor_move_base_nodate}. All simulation tests are run one at a time in the Arena-RosNav simulation environment on a Dell Precision 3660 Tower with an Intel i9-12900K CPU with 16 cores (single-thread passmark of $4,336$; multi-thread score of $41,322$). All hardware tests are run on a Dell XPS 13 laptop with an Intel i5-10210U CPU with $4$ cores (single-thread passmark of $2,145$; multi-thread score of $6,152$).

\subsection{Experiment One: Assumption-satisfying Experiments}
\label{sec:experiment_1}

\begin{figure}[h!]
    \centering
    \includegraphics[width=0.98\linewidth]{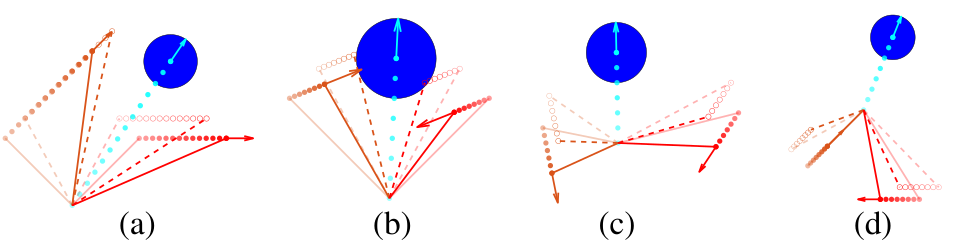}
    \caption{Visualization of four Monte Carlo variations of gaps from Experiment One. Orange points and lines represent the left side of the gap while red points and lines represent the right side. Solid points and lines represent the original gap geometry whereas hollow points and dashed lines correspond to the inflated version of the gap. Transparent points represent positions at prior timesteps. The blue circle represents the robot along with its finite radius.}
    \label{fig:matlab_benchmark}
\end{figure}

First, example trajectories are generated through single gaps to demonstrate how under ideal conditions, pursuit guidance-based policies can generate provably safe trajectories. A single gap is randomly generated and the PN policy is employed for trajectory generation. With the gap centered at the origin, left gap points are uniformly sampled from $\beta_l \in [\frac{\pi}{2}, \frac{3 \pi}{2}]$, $r_l \in [0.25, 1.0]$ m. Right gap points are uniformly sampled from $\beta_r \in [\frac{-\pi}{2}, \frac{\pi}{2}]$, $r_r \in [0.25, 1.0]$ m. Gap point velocities are uniformly sampled from all directions with magnitudes within the range $[0.0, 1.0]$ m/s. A finite robot radius of $0.20$~m is also accounted for by the inflation policy in Section \ref{sec:gap_point_inflation}. A subset of gaps are shown in Figure \ref{fig:matlab_benchmark}.

For this experiment, $10,000$ trials were run: $6,987$ trials end in collision-free gap passage, $2,668$ trials resulted in a kinematically infeasible gap due to the velocity limits of the robot, and for the remaining $345$ trials, the robot was unable to pass through the gap before it closed. No collisions occurred during any trials.

\subsection{Experiment Two: Canonical Scenarios}
\label{sec:experiment_2}

In this section, a collection of isolated canonical scenarios are demonstrated to provide further insights into how dynamic gap plans and predicts in dynamic settings.

\subsubsection{Closing and Re-opening}
\label{sec:closing_and_reopening}

\begin{figure}[h!]
    \centering
    \includegraphics[width=0.98\linewidth]{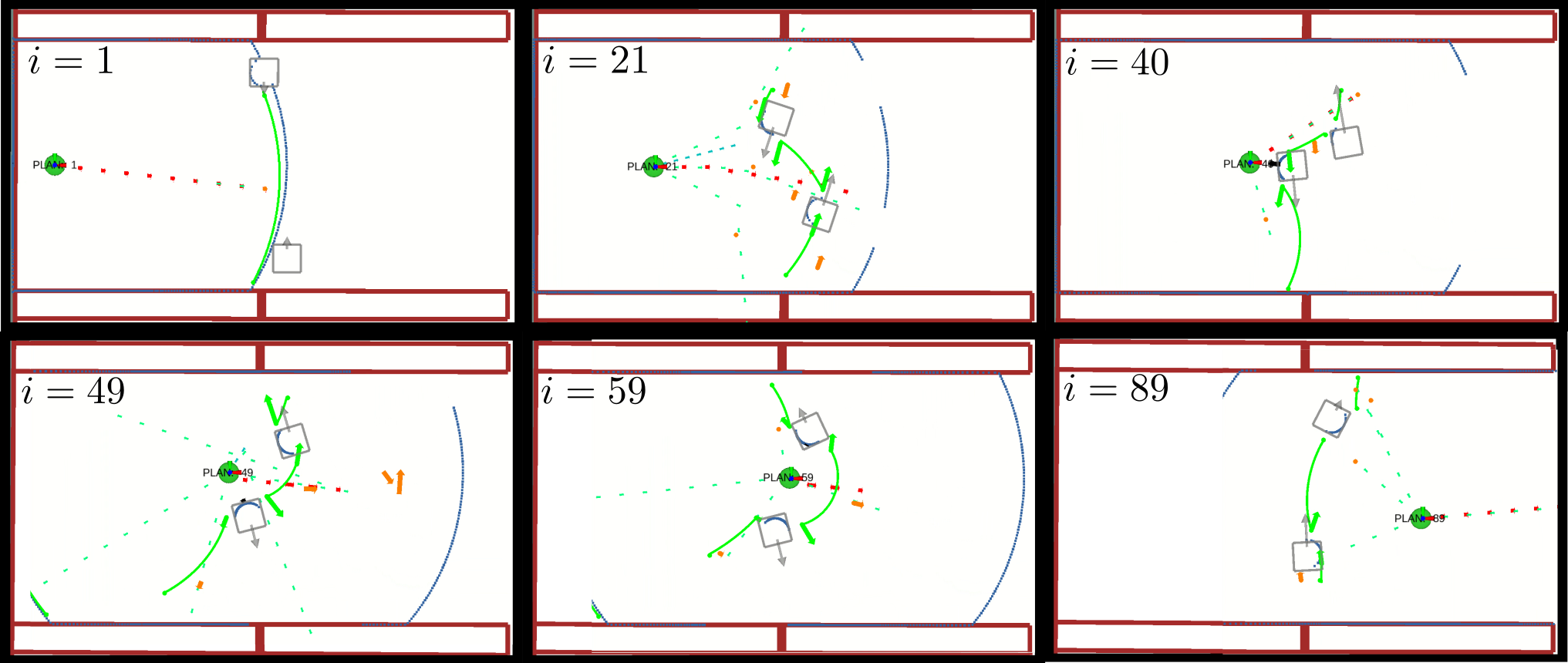}
    \caption{Canonical Scenario One - closing and re-opening gap. The green circle with the attached frame is the ego-robot, and the gray squares are the agents. The dotted blue arc shows the current scan data, green arcs show the current manipulated gaps, orange arrows represent the position and velocity of the local gap goals, light blue pointed lines are the generated trajectories, and the red pointed line is the trajectory that is currently being executed by the robot.}
    \label{fig:cs1_reopening}
\end{figure}

In this scenario (Figure \ref{fig:cs1_reopening}), the ability of dynamic gap to predict gaps to close and subsequently re-open is highlighted. In this scenario, the robot must move from the left side of the corridor to the right side beyond the two agents. The frame notations $i=n$ in Figure \ref{fig:cs1_reopening} refer to the planning loop iteration at which the frame was captured. At $i=21$, the planner can propagate the central gap forward in time to determine that it will close and re-open during the planning horizon. Therefore, a piecewise trajectory is generated that leads the ego-robot up to the crossing gap, holds the ego-robot in place while the gap reopens ($i=40$), and then leads the ego-robot through the newly opened gap ($i=49$). % The frames $i=59$ and $i=89$ show the robot planning through the reopened gap.

% At $i=1$, planning begins but the scan is not large enough to detect the agents. Therefore, a central trajectory is generated through the closing gap. 
%  to reach the other end of the corridor

The original version of dynamic gap only propagated gaps forward up until the point at which their dynamics are altered, in this scenario being when the central gap closes. This means that the original dynamic gap planner has no way to predict that the central gap will reopen in the future for safe gap passage.

\subsubsection{Corridor - receding}
\label{sec:corridor_overtaking}

\begin{figure}[h!]
    \subfloat[Canonical Scenario Two - receding corridor.]{\includegraphics[width=0.59\linewidth]{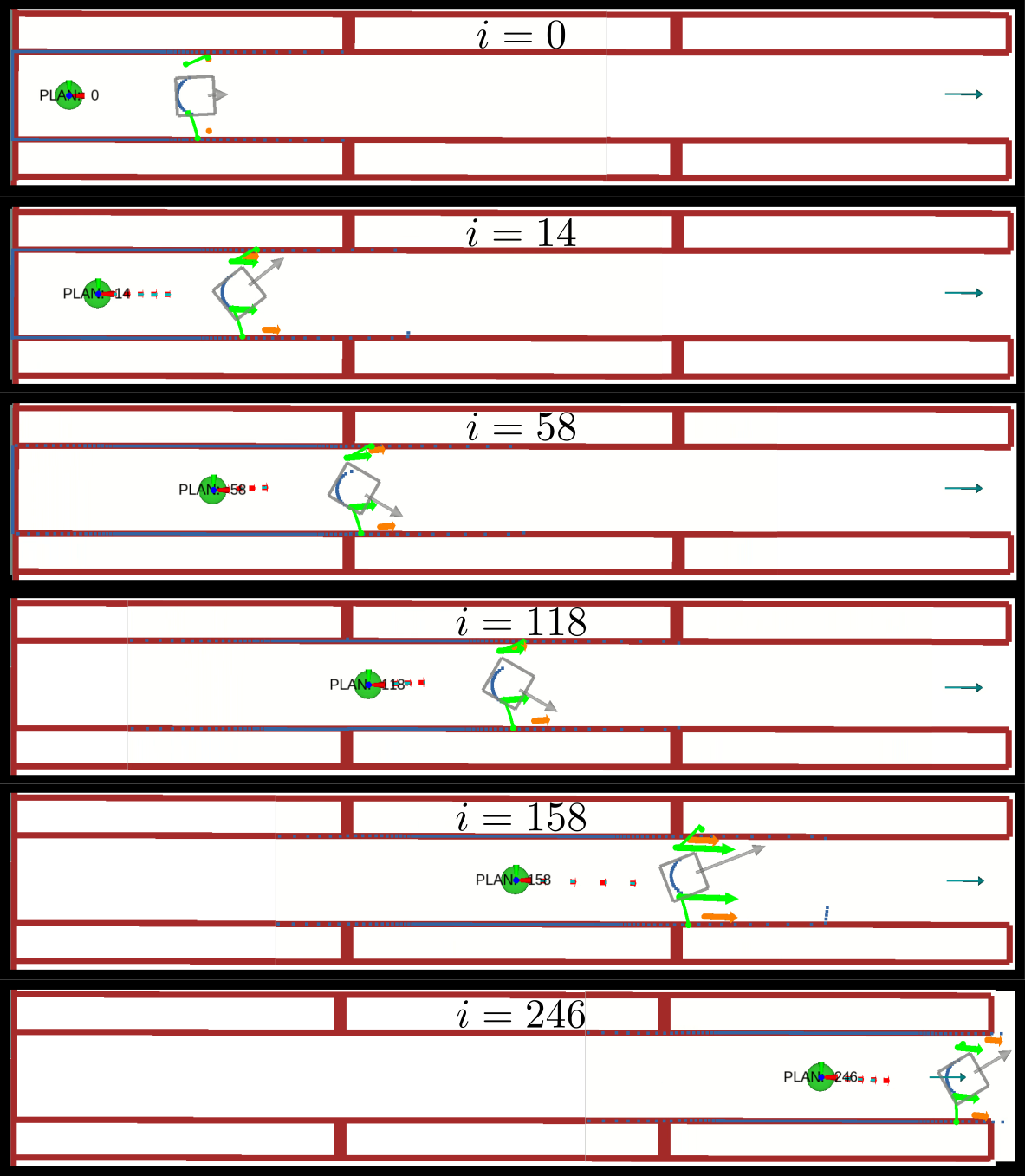}\label{fig:cs2_overtaking}}
    \quad
    \subfloat[Canonical Scenario Three - approaching corridor.]{\includegraphics[width=0.39\linewidth]{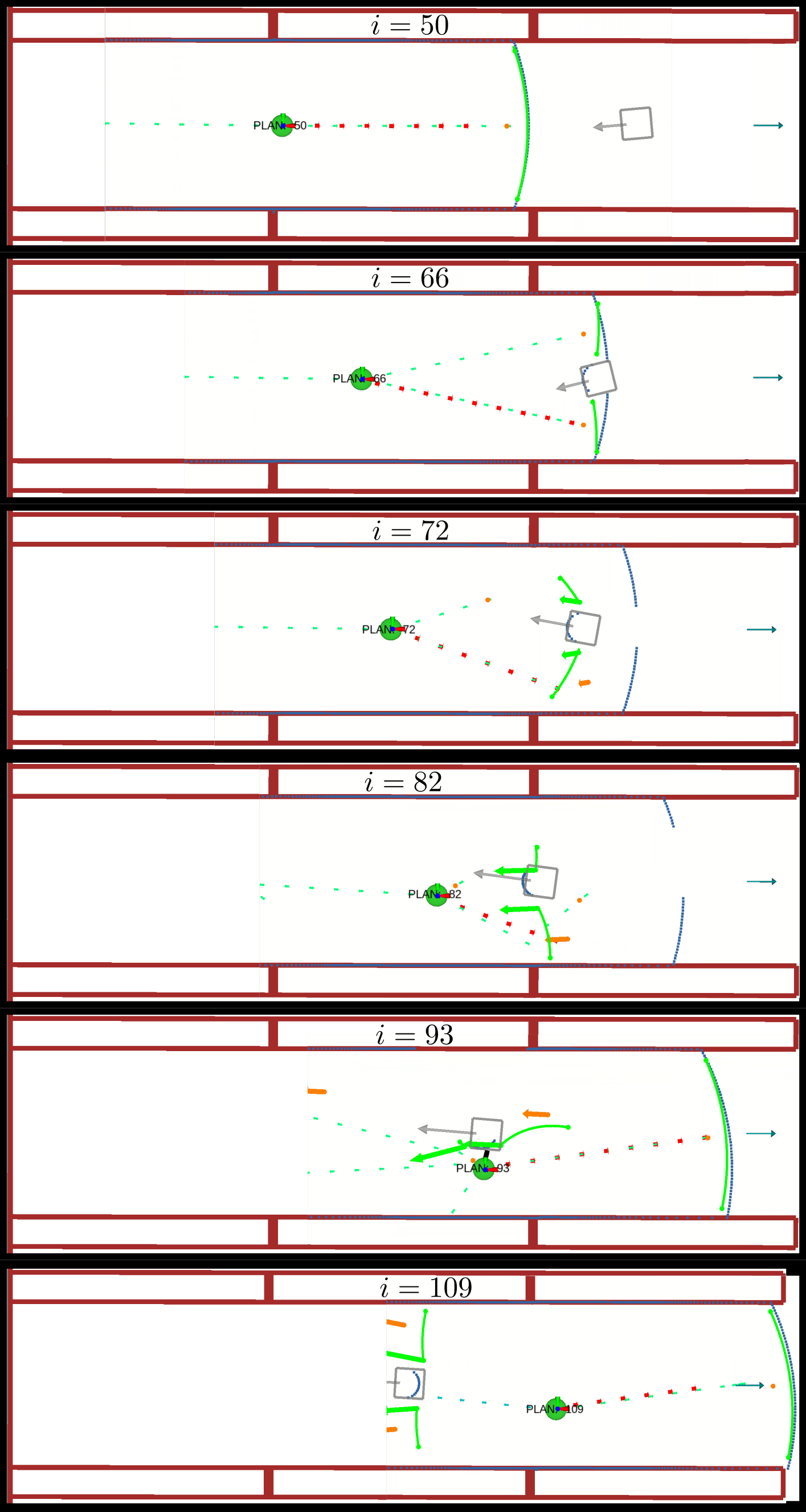}\label{fig:cs3_approaching}}
    \caption{Canonical Scenarios Two and Three - corridor.}
\end{figure}

In this scenario (Figure \ref{fig:cs2_overtaking}), the ability of dynamic gap to plan through a presently occupied region of space, referred to as the \say{ungap}, is highlighted. In this scenario, the robot must move from the left side of the corridor to the right side, where the goal is represented by the blue arrow. In this setup, the corridor is deliberately designed to be too narrow for the ego-robot to move around the agent to pass it and continue on. There are two gaps detected on either side of the agent, but the trajectories are short and would cause collisions with the walls. Starting at frame $i=14$, the planner can be seen to select the central ungap trajectory that allows the ego-robot to trail behind the agent. This ungap trajectory is consistently generated and tracked in all subsequent frames up until $i=246$ when the robot reaches the goal. 

%Therefore, the only way to move through this corridor is to plan in the space that the agent is presently occupying. 
 
The original dynamic planner could not generate trajectories through ungaps. Therefore, the planner would have to wait for the agent to clear out of the nearby space before it could plan a trajectory through the corridor.

\begin{comment}
\begin{figure}[h!]
    \centering
    \includegraphics[width=0.75\linewidth]{figures/cs2.png}
    \caption{Canonical Scenario 2 - receding corridor.  The green circle with the attached frame is the ego-robot, and the gray squares are the agents. The dotted blue arc shows the current scan data, green arcs show the current manipulated gaps, orange arrow represent the position and velocity of the local gap goals, dark blue pointed lines are the generated trajectories through the ungap, and the red pointed line is the trajectory that is currently being executed by the robot.}
    \label{fig:cs2_overtaking}
\end{figure}
\end{comment}

% \newpage

\subsubsection{Corridor - approaching}
\label{sec:corridor_approaching}

In this scenario (Figure \ref{fig:cs3_approaching}), the ability of dynamic gap to plan around an oncoming obstacle in a tight corridor is demonstrated. Similar to the last scenario, the planner must get the robot from the left side of the corridor to the right side, where the goal is represented by the blue arrow. In this setup, the corridor is wider compared to prior scenario, thereby allowing for the ego-robot to move around the agent to pass it and continue on.  At iteration $i=66$, the oncoming obstacle has been detected in the laser scan, but the gap point models have yet to converge to an accurate velocity estimate. By iteration $i=72$, the approaching agent is both detected and predicted to be traveling towards the ego-robot. The planner successfully predicts this agent forward, generating two trajectories that pass the agent on the left and right sides.

% , ultimately selecting the right
% At planning iterations $i=50$, the sensing radius is not large enough to detect the oncoming agent, leading the planner to simply plan a trajectory straight forward.
% The planner successfully maneuvers around the agent in iterations $i=82$ and $i=93$ and continues planning to the goal by iteration $i=109$. 

\begin{comment}
\begin{figure}[h!]
    \centering
    \includegraphics[width=0.6\linewidth]{figures/cs3.png}
    \caption{Canonical Scenario 3 - approaching corridor.  The green circle with the attached frame is the ego-robot, and the gray squares are the agents. The dotted blue arc shows the current scan data, green arcs show the current manipulated gaps, orange arrow represent the position and velocity of the local gap goals, light blue pointed lines are the generated trajectories, and the red pointed line is the trajectory that is currently being executed by the robot.}
    \label{fig:cs3_approaching}
\end{figure}
\end{comment}
% \newpage

\subsubsection{Four-way intersection}
\label{sec:four_way_intersection}

\begin{figure}[h!]
    \centering
    \includegraphics[width=0.75\linewidth]{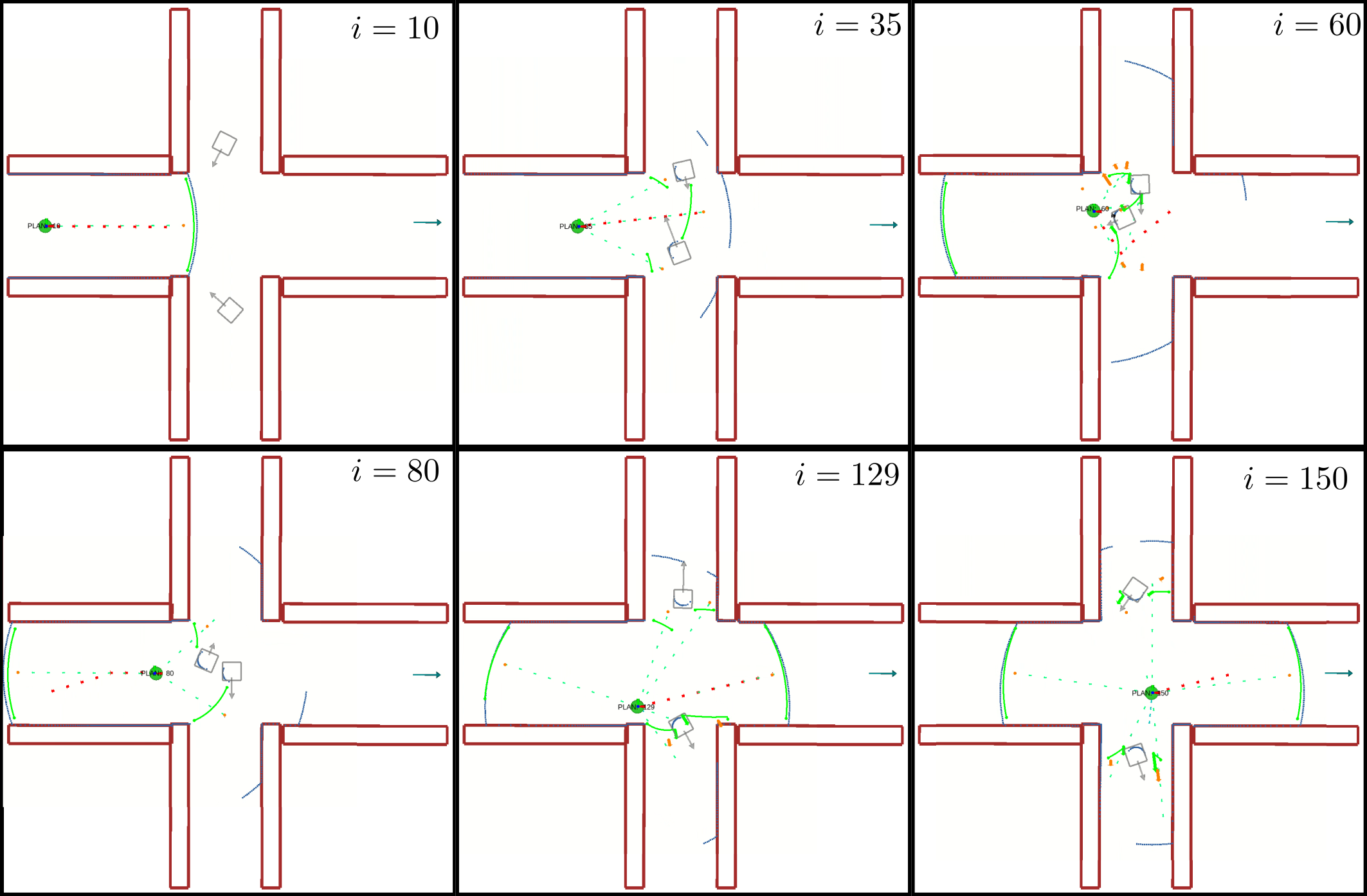}
    \caption{Canonical Scenario Four - intersection.}
    \label{fig:cs4_intersection}
\end{figure}

This scenario (Figure \ref{fig:cs4_intersection}) highlights the ability of the dynamic gap planner to plan through a complex four-way intersection scenario. In this scenario, the planner must move the robot from the left branch of the intersection to the right branch, where the goal is represented by the blue arrow. In this setup, two agents pass through the top and bottom branches of the intersection. 

By iteration $i=35$, the agents are detected and the planner initially opts to pass between the two agents. However, at planning iteration $i=60$ the two agents are attempting to de-conflict their own motion plans and staying in the center of the intersection, making it difficult for the planner to pass through them. Due to this, the planner moves the ego-robot away from the intersection at iteration $i=80$. By iteration $i=129$, the gap between the agents has re-opened and the planner generates a trajectory through it to continue through the intersection.

% At planning iteration $i=10$, the sensing radius is not large enough to detect the agents, leading the planner to simply plan a trajectory straight forward. 
%  At iteration $i=150$, the agents turn around to pass through the intersection a second time, but at this point the planner has generated a trajectory that will allow it to pass through the intersection.

\subsection{Experiment Three: Simulation Benchmarking}
\label{sec:simulation_benchmarking}

To gain a better understanding of the performance level of the dynamic gap planner, a comprehensive series of simulation benchmarks are run with other state-of-the-art local perception-informed planners. The dynamic gap planner is integrated into the Arena-Rosnav \cite{kastner_arena-rosnav_2021} benchmarking environment and compared against four classical cost map-based planners and  four learned planners. Additionally, two other GBPs are tested against including the static world predecessor of dynamic gap, known as potential gap, and the prior version of dynamic gap. The version of dynamic gap proposed in this chapter is benchmarked under both a holonomic robot model as well as a nonholonomic robot model. These baselines provide insights into the performance improvements obtained from the work presented in this chapter. All baselines are detailed in Section \ref{sec:baselines}.

\begin{figure}[h!]
    \centering
    \includegraphics[width=0.99\linewidth]{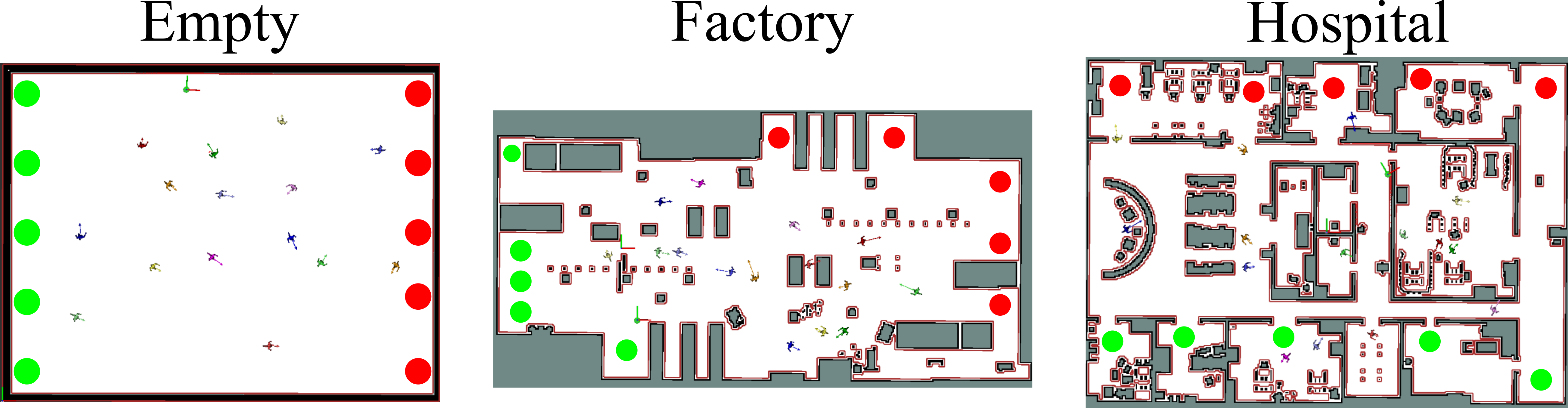}
    \caption{Three simulation environments used during benchmarking. Sizes are Empty - $31 \times 24$ m, Factory - $38 \times 19$ m, Hospital - $34 \times 24$ m. Green dots represent the different start positions and red dots represent different goal positions for simulation benchmarking. Each combination is evaluated one time, yielding $5 \times 5 = 25$ trials for each environment.}
    \label{fig:arena_environments}
\end{figure}

Three environments are used, shown in Figure \ref{fig:arena_environments}, referred to as empty, factory, and hospital. With these environments, the authors aim to build a gradient of increasing environment structure to evaluate each planner's ability to not only navigate dynamic obstacles, in simulation represented as pedestrians, but also the static, non-trivial structure of these worlds such as corridors, rooms, and atria.

Within each environment, 15 dynamic agents are placed at predefined start points. Then, a path is generated from each agent's start to its goal which the agent subsequently tracks. Both the agents and the ego-robot have velocity limits of $v_x^{\rm max}=v_y^{\rm max}=\omega_z^{\rm max}=1.0$ m/s. For each planner / environment combination, 25 tests are run, consisting of every combination of the five start and goal positions portrayed in Figure \ref{fig:arena_environments}. These $25$ trials are also run with no agents present to give insight into how the planners perform in a purely static version of each environment. For all trials, planners are given three minutes to reach the goal. 

% These agents do not have collision avoidance methods; they instead perform open-loop path tracking with noise added to their velocities. 

For navigation-level performance, two dependent variables are captured: whether or not the planner reached the goal under the prescribed time limit for the given trial, and whether or not the planner registered any collisions during the trial. Trials in which the planner reached the goal under the time limit without sustaining any collisions are considered a \textit{success}. Trials in which the planner does not reach the goal before the time limit, but also does not sustain any collisions are depicted as \textit{timed out}. Trials for which the planner reaches the goal under the time limit but sustains collisions are treated as \textit{failed}. Trials that both did not reach the goal before the time limit and also sustained collisions are depicted as \textit{failed and timed out}. Results are reported by environment in Sections \ref{sec:empty}, \ref{sec:factory}, and  \ref{sec:hospital}.

\subsubsection{State-of-the-art Baselines}
\label{sec:baselines}

In this chapter, four classical cost map-based planners are evaluated:

\textbf{Timed Elastic Bands (TEB) \cite{rosmann_timed-elastic-bands_2015}:} TEB is a local planner which formulates a soft constraint optimal control problem with respect to trajectory execution time, obstacle separation, and compliance with kinodynamic constraints. The planner performs blob detection on a local costmap to detect obstacles and runs KFs to estimate the dynamic obstacle states.

% The optimization formulation applies penalty methods to convert the original problem into an unconstrained optimization problem. This optimal control problem is solved internally with g2o \cite{kummerle_g2o_2011}. 

\textbf{Co-operative Human-Aware Navigation (CoHAN) \cite{teja_singamaneni_human-aware_2021}:}
CoHAN is a tunable human-aware navigation planner designed to handle diverse human-robot interaction contexts through context-based planning modes and a social compliance cost function. From a planning perspective, CoHAN is a socially aware version of TEB. 

\textbf{Model Predictive Control (MPC) \cite{rosmann_online_2021}:} The authors from TEB also developed a receding horizon-based optimal controller that solves an optimal control problem much like the one formulated in the original TEB work.

\textbf{Dynamic Window Approach (DWA) \cite{fox_dynamic_1997}:} DWA randomly samples the robot's control space ($v_x$, $v_y$, $\omega$) and rolls out the resultant trajectory for a short period of time. Each candidate trajectory is evaluated on proximity to obstacles, the global goal, and the global path as well as speed. Colliding trajectories are discarded, but the implementation that is benchmarked against has no dynamic obstacle prediction.

In addition, four data-driven planners are evaluated:

\textbf{Deep Reinforcement Learning - Velocity Obstacles (DRL-VO) \cite{xie_drl-vo_2023}:}
DRL-VO is a deep reinforcement learning approach with a reward function based on VOs that penalizes headings that are likely to lead to collisions using relative motion rather than distance. Inputs include pedestrian states, a sliding window of laser scans, and a local waypoint. Their policy is trained with the PPO (Proximal Policy Optimization) algorithm in the Gazebo simulation. 

% that uses  a combination of lidar history, pedestrian kinematics, and a sub-goal point. It uses 
% which is mapped to a 2D pedestrian map
%  that are stacked and pooled into a Lidar map
% The policy outputs a linear and angular velocity command. 
\textbf{Reinforcement Learning for Collision Avoidance (RLCA) \cite{long_towards_2018}:}
The RLCA method is a reinforcement learning model that directly maps the raw sensor data, in this case a laser scan, to a desired, collision-free steering command. For this specific network, a sliding window of the three past scans are passed into the model along with a local waypoint from the global path and the robot's current velocity. The reward function for this model is comprised of three terms. First, a reward term for making progress towards the local waypoint. Second, a flat penalty for collisions. Lastly, a regularization term to discourage large angular velocities. This formulation is solved using PPO in the Stage simulator \cite{vaughan_massively_2008}.

% are input and passed through a series of 1D convolutions
% Outputs are then linear and angular velocity values that are sent to the robot.
% , and otherwise administers small rewards for making forward progress towards the goal while penalizing moving away from the goal.
% is administered if the ego-robot collides with any static or dynamic obstacles in the environments
% is added
\textbf{Collision Avoidance with Deep Reinforcement Learning (CADRL) \cite{everett_motion_2018}:}
For CADRL, ground truth state information for nearby agents, ---including position, velocity, and radius --- is passed into model through a Long Short-Term Memory (LSTM) network, and the ego-robot state information --- including a local waypoint from the global path, the preferred velocity, and the orientation --- is then appended to the input. The reward function administers a positive value if the local waypoint is reached, a linearly decreasing negative value if an agent is close to the robot, and a flat negative value if an agent is in collision with the robot. This model is trained with Asynchronous Advantage Actor-Critic (A3C) algorithm using a custom simulation engine. 

\textbf{Learning from Learned Hallucination (LfLH) \cite{wang_agile_2021}:}
LfLH is a self-supervised framework that leverages automated hallucination to generate synthetic obstacle configurations during training rather than relying on trial-and-error exploration. An encoder-decoder structure is used for hallucination in which the robot state, goal state, and robot plan are passed to the encoder during training. The encoder then learns to predict probability distributions of nearby obstacles. The decoder samples from these obstacle distributions and generates a motion plan. Then, a motion planner is learned through behavior cloning by sampling the learned encoder to generate hallucinated obstacle perceptions and comparing the output action against the actions of an optimization-based motion planner that is treated as an expert under this framework.

% Their approach also leverages classical techniques including model predictive control for additional safety. 

Lastly, a family of GBPs are evaluated:
% to properly contextualize the proposed dynamic gap planner

\textbf{Potential Gap (PGap) \cite{xu_potential_2021}:}
Potential Gap is a gap-based planner designed for static environments. This planner detects the current set of gaps in the local environment, manipulates the set of gaps to ensure gaps are convex polar triangles, and then deploys an attractive potential and circulation field to drive the robot through gaps. 

\textbf{Dynamic Gap (DGap) \cite{asselmeier_dynamic_2025}:} This baseline is the prior version of the planner discussed in this chapter. This baseline propagates gaps only up until the point in which they close or propagate into another gap. This baseline also has no means to generate trajectories through ungaps.

\subsubsection{Empty}
\label{sec:empty}
\begin{figure}[h!]
    \centering
    \includegraphics[width=0.99\linewidth]{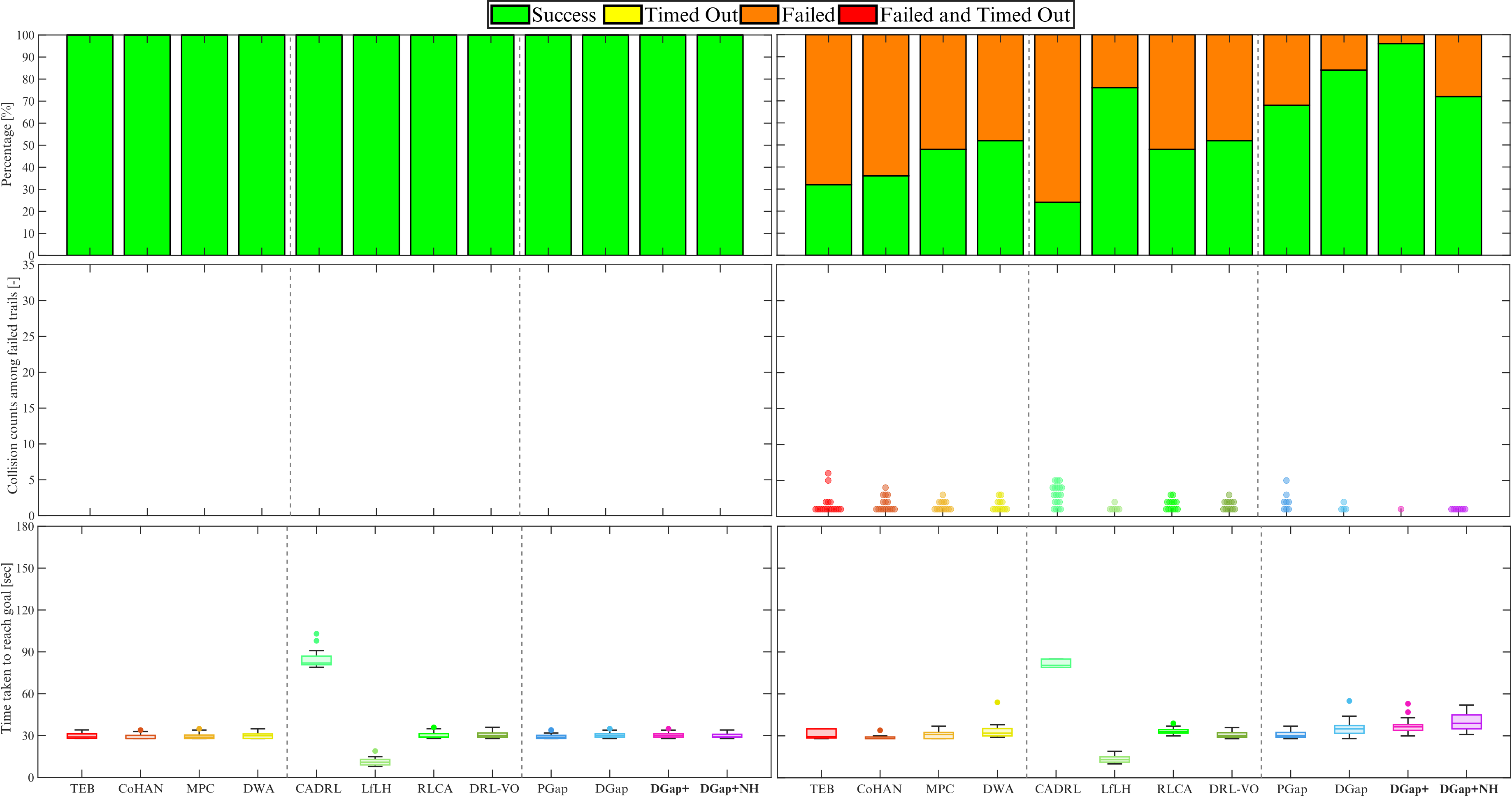}
    \caption{Simulation benchmarking results for the Empty simulation environment. The left column represents trials with no agents in the environment, and the right column is for trials with 15 agents in the environment. The top row portrays navigation performance results for all benchmarks across the 25 trials. Green bars (success) are successful trials in which the planner reached the goal under the time limit without sustaining any collisions, yellow bars (time out) are trials for which the planner does not reach the goal before the time limit but also does not sustain any collisions, orange bars (failure) are trials for which the planner reaches the goal under the time limit but sustains collisions, and red bars (failure and time out) are for trials that do not reach the goal under the time limit and also sustain collisions. Results are shown in percentage format. The middle row depicts the collision counts amongst failed trials, meaning how many collisions actually occurred during each trial that registered a positive number of collisions. Lastly, the bottom row depicts average time taken to reach the goal for successful trials.}
    \label{fig:empty_combined}
\end{figure}

Given that this environment is solely comprised of a single room, these trials can be used to evaluate each planner's pure dynamic obstacle avoidance mechanisms. Results for this environment are shown in Figure \ref{fig:empty_combined}. For the static version of the empty environment, all planners succeeded on all $25$ trials. The only notable takeaway from this set of tests is that the CADRL planner took roughly three times longer ($79 - 91$ seconds) to reach the goal compared to other baselines ($28 - 36$ seconds).

%  This is the simplest navigation task possible, where there are no obstacles and the shortest feasible path from start to goal is simply a straight line. 
%  some of the learned planners demonstrate quite different times to reach the goal than the normal times for other baselines. Specifically, 
%  The LfLH planner took roughly a third to a half as long ($8 - 18.96$ seconds). 

For the dynamic version of the empty environment, failure modes start to arise. In this setting, all unsuccessful trials are the result of collisions, and no timeouts were observed. Average performance across the classical cost map-based planners (the leftmost four benchmarks) is on par with the average performance for the learned planners (the middle four benchmarks), with the classical planners averaging a success rate of $42\%$ while the learned planners average a success rate of $50\%$. However, the learned planners exhibit more variance in that CADRL performs worse ($24\%$ success rate) and LfLH performs better ($76\%$ success rate) compared to the average. Of the three environments, this is the one in which the set of learned planners performs the best. This is in part due to the environment closely resembling the training environments used to optimize the learned models, meaning that the scenarios encountered in this environment were more likely to exist inside of the distribution of data used to train the models.

%  This is sensible given that the environment layout is so simple. 
% Insert a bit on the classical planners
While the costmap representation provides a way to discretely identify obstacles, this set of planners still struggled at times to avoid oncoming obstacles. These planners were often able to slightly adjust their heading to direct the ego-robot away from nearby agents. However, when agents entered the local costmap heading towards the ego-robot, the planners were not able to react in time to avoid the agent.

% Insert a bit on the learned planners
For the learned benchmarks, the planners all possessed the emergent property of coming to a stop directly in front of obstacles when the robot encroaches upon them, relying on non-adversarial motions of nearby agents to avoid collisions. 

% Insert a bit on the gap-based planners
The GBPs all performed well in this setting. Gaps proved to be an effective representation due to their ability to naturally bias the planning space away from nearby obstacles. When agents did approach the ego-robot, the gap detection policy naturally partitions the planning space to either side of these agents which guides planning around the agents. The proposed version of dynamic gap deployed on a holonomic robot model outperforms all other baselines in this environment.

\begin{comment}

\begin{figure}[h!]
    \centering
    \includegraphics[width=0.99\linewidth]{figures/empty_static.png}
    \caption{Empty environment, no agents.}
    \label{fig:empty_static}
\end{figure}

\begin{figure}[h!]
    \centering
    \includegraphics[width=0.99\linewidth]{figures/empty_dynamic.png}
    \caption{Empty environment, 15 agents.}
    \label{fig:empty_dynamic}
\end{figure}
\end{comment}
% \newpage

\subsubsection{Factory}
\label{sec:factory}

\begin{figure}[h!]
    \centering
    \includegraphics[width=0.99\linewidth]{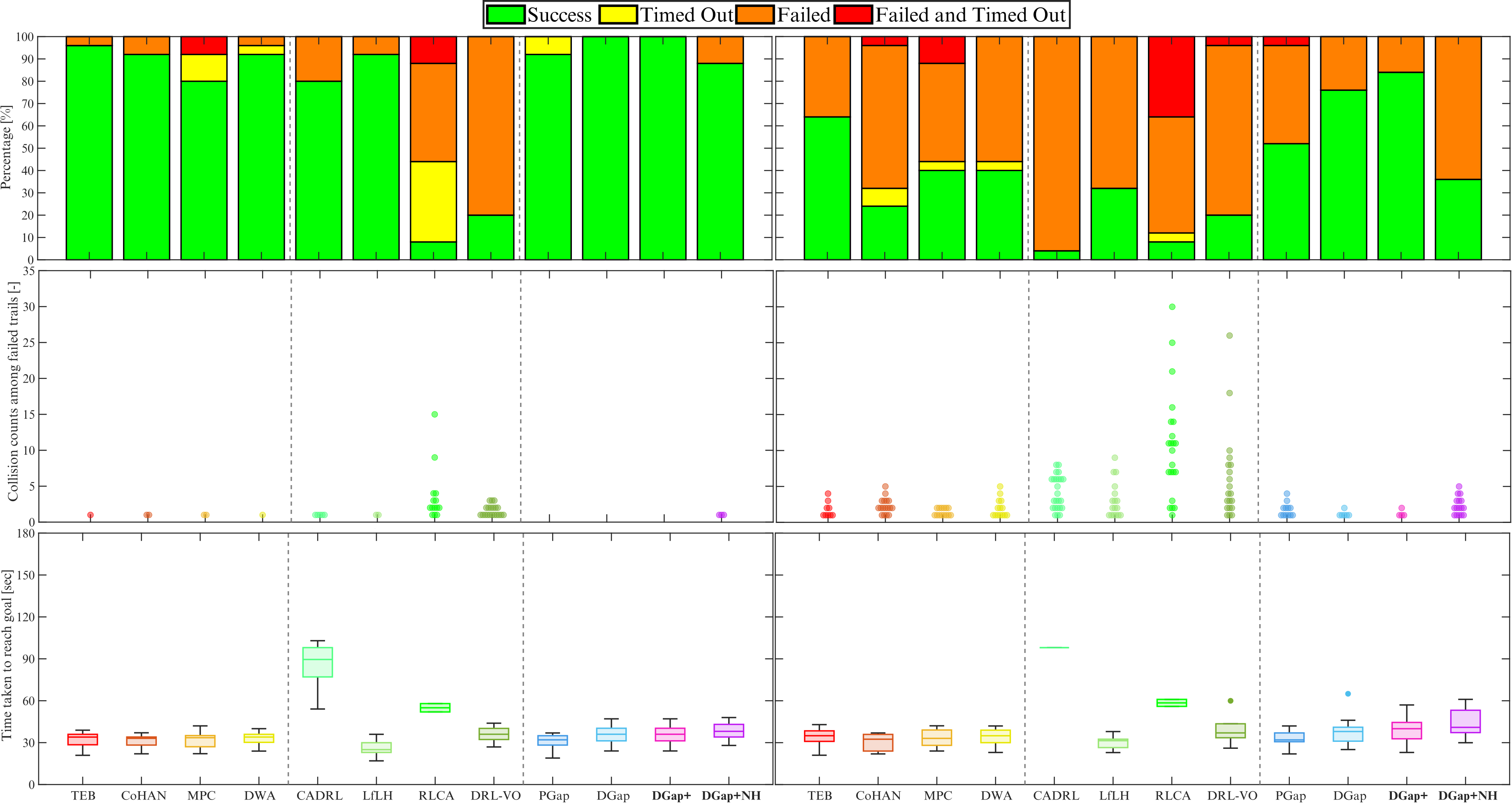}
    \caption{Simulation benchmarking results for the Factory simulation environment.}
    \label{fig:factory_combined}
\end{figure}

The factory setting consists of one large room with many smaller isolated static obstacles positioned within it, resembling real-world artifacts such as tables or pillars. The results for this environment are shown in Figure \ref{fig:factory_combined}. Here, the larger regions of free space allow multiple agents to pass through the same part of the environment at one time. This structure causes more collisions than the empty environment. 

The four learned baselines all accept either a single raw laser scan or a sliding window of raw laser scans as their sensory input, meaning that these planners do not perform any form of environment abstraction. Building an environment representation can serve many purposes: reducing input size by distilling large sensor data into lower-dimensional formats, extracting additional unobservable features from the sensor data, and preprocessing the noisy sensor data to remove artifacts or inaccuracies. By foregoing this environment abstraction step, the task of extracting meaningful environment information from inputs becomes much more difficult for a learned model. The authors posit that this is one of the reasons for the significant performance drop amongst learned planners. The RLCA and DRL-VO models both struggled heavily with the static structure of the environment and timed out on a handful of trials because of their inability to get through narrow corridors or around tight corners. These planners also rarely backed up when in collision with the environment which can often help get the robot reset. The CADRL planner was immensely slow, but this ended up helping the planner navigate through the static environment. The other learned planners often over-rotated or over-compensated other maneuvers, but CADRL was instead very deliberate. This allowed it to reach the goal consistently, but at the cost of very limited collision avoidance abilities due to its slow pace. Not only do the learned planners exhibit more failed trials, the individual failed trials also sustain a far greater number of collisions. % During these trials, the learned planners do not appear to attempt significant collision avoidance maneuvers such as swerving to avoid an oncoming agent. 

%  in moving through the environment

% The factory environment provides an interesting platform to discuss the trade-offs of the different environment representations used among these baselines. 

The four classical benchmarks all use a local costmap that is propagated through the environment as the robot moves. This costmap takes in scan data and updates a discrete occupancy grid. Then, this occupancy grid is inflated, scored, and planned over. The Cartesian frame costmap is a very simple and intuitive environment representation that makes planning algorithms such as search and optimization easy to execute. Some baselines also perform motion estimation steps on this costmap to predict how the map might evolve in the future. The classical baselines perform better in the Factory environment than the learned baselines with no agents present as well as with 15 agents present. However, the classical planners occasionally clipped corners or hallways during navigation. Without agents, the classical baselines register success rates of $80-96\%$ compared to the success rates of $8-92\%$ from the learned baselines. With 15 agents, the classical baselines register success rates of $24-64\%$ compared to the $4-32\%$ success rates of the learned baselines.

% This costmap captures the local environment structure as well as how the robot geometry might interact with this structure, discounting paths that robot can not fit in or reach. 

% \begin{itemize}
    % \item Learned methods register far more collisions
    % \item All learned methods are acting on raw laser scan
    % \item Raw laser scan is an insufficient environment abstraction for collision avoidance
% \end{itemize}

The three gap-based benchmarks all plan using a gap-based environment representation comprised of polar arcs of free space. This environment representation requires minimal computation to synthesize and keeps the scan data in its raw polar format, bypassing the preprocessing steps required to maintain a costmap. Being able to build and propagate this gap-based representation pays dividends in practice, with the family of GBPs exhibiting the best navigation performance of all baselines with success rates of $36-84\%$ in the dynamic setting. The ability of dynamic gap to predict the feasibility of local gaps before committing to them proves paramount in avoiding collisions in this environment. The proposed version of the dynamic gap planner on the holonomic model outperformed the rest of the baselines with its added gap propagation abilities, although the proposed planner on the nonholonomic model only achieved a success rate of $36\%$. This can largely be attributed to the model mismatch in which trajectories are planned through pursuit guidance techniques under a holonomic model, but then tracked under a nonholonomic model. The focus of this work was planning in dynamic environments, and future work will extend dynamic gap to a nonholonomic model. 

\begin{comment}
\begin{figure}[h!]
    \centering
    \includegraphics[width=0.99\linewidth]{figures/factory_static.png}
    \caption{Factory environment, no agents.}
    \label{fig:factory_static}
\end{figure}

\begin{figure}[h!]
    \centering
    \includegraphics[width=0.99\linewidth]{figures/factory_dynamic.png}
    \caption{Factory environment, 15 agents.}
    \label{fig:factory_dynamic}
\end{figure}
\end{comment}

\subsubsection{Hospital}
\label{sec:hospital}

\begin{figure}[h!]
    \centering
    \includegraphics[width=0.99\linewidth]{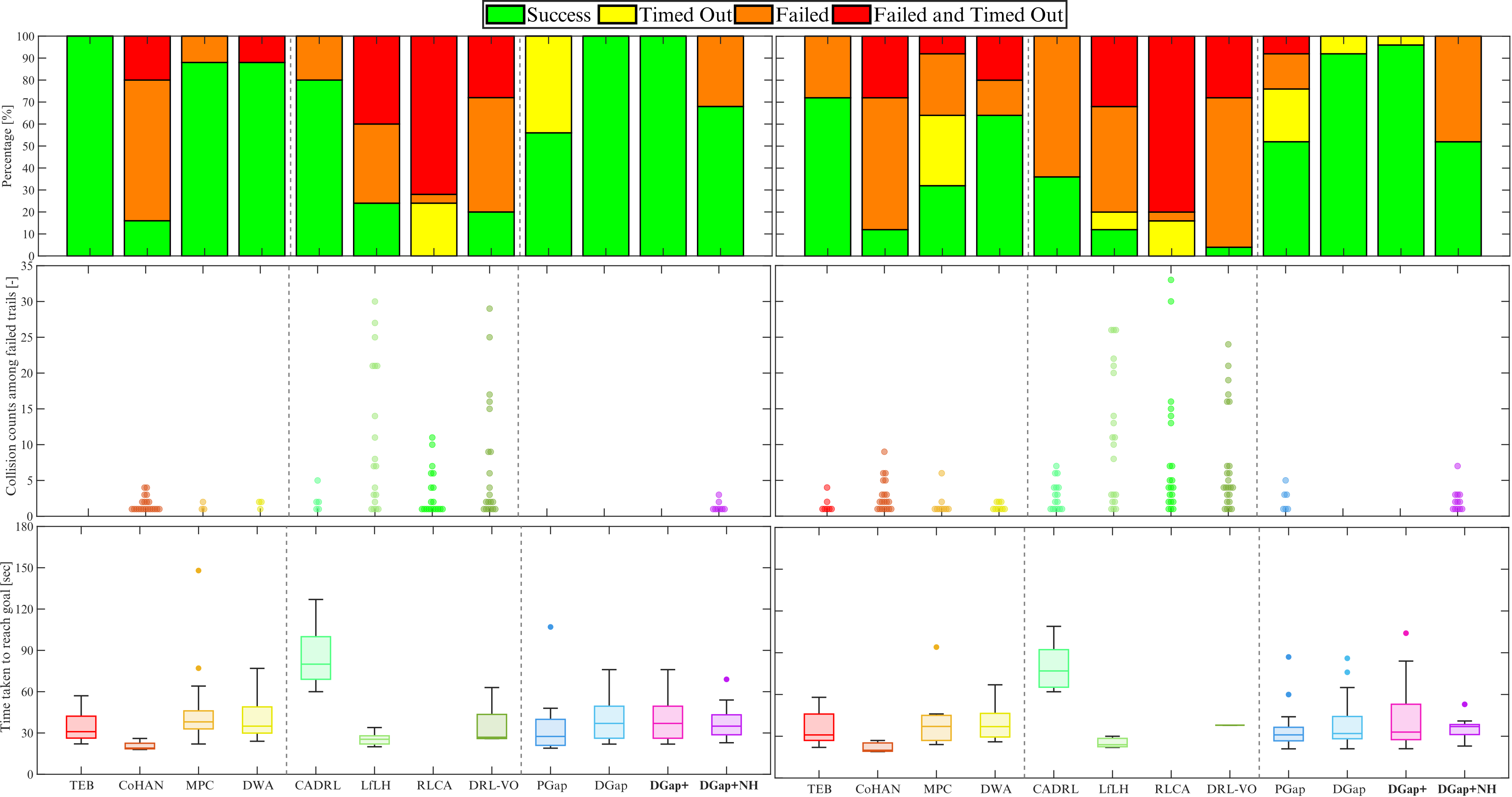}
    \caption{Simulation benchmarking results for the Hospital simulation environment.}
    \label{fig:hospital_combined}
\end{figure}

The hospital environment exhibits many smaller rooms connected through tighter passageways and corridors. Results for this environment are shown in Figure \ref{fig:hospital_combined}. Compared to the factory environment, fewer collisions are registered in this environment because the corridors and small rooms offered fewer opportunities for several agents to enter the same region and collide. Here, more trials ended in time outs, either due to the planner getting stuck in a tight space and not being able to make further progress in the environment, or due to the planner taking too long to reach the goal. Given that this environment is more sprawling, the time taken to reach the goal is higher on average, and more variance is seen in navigation times.

% The trends in collisions from the factory environment results continue in the hospital environment. 
The CoHAN baseline tended to rotate and translate while trying to pass around corners. As a result, the planner would often bump into the corner and \say{roll} around it, resulting in several collisions. The MPC and DWA baselines occasionally drove backwards through corridors as well.

The learned planners exhibited more failed trials and higher collision counts. Overall, learned planners struggled with planning paths through narrow corners and corridors. The RLCA planner tended to over-rotate, leading the ego-robot to hit the walls surrounding corridor entrypoints. Once the robot hit the wall, the planner was unable to output negative velocity commands to have the robot back up, and trial progress would typically end. The DRL-VO planner exhibited a strong bias in its output space distribution, solely outputting positive (counterclockwise) angular velocities. This would lead to situations where a simple hallway corner that could be passed through with a clockwise rotation could take very long due to the planner's inability to rotate in that direction. The authors mitigated this bias by shifting the interval of possible angular velocities towards negative values, but this only provided marginal performance improvements. The CADRL baseline would commonly output extremely small velocity values when the robot was close to static obstacles. This led to situations in which the planner would take a corner too tight and stall out next to the wall. The authors partially alleviated this issue by applying a small constant forward velocity to the system when there were no obstacles in front of the robot, enabling the robot to drive through these scenarios where the planner stalls out. The CADRL and LfLH planners both possessed a \say{rotate-in-place} mode that would guide the planner in aligning the ego-robot to point towards the next waypoint from the global goal. This helped these planners get through difficult portions of the environment.

Overall, the GBPs fared well in this setting. While not designed for dynamic settings, the planning behaviors of the potential gap planner still allowed it to navigate through this environment's internal structure. The two dynamic gap versions performed very well, with the holonomic proposed version narrowly outperforming its prior version with a success rate of $96\%$ versus $92\%$.
\begin{comment}
\begin{figure}[h!]
    \centering
    \includegraphics[width=0.99\linewidth]{figures/hospital_static.png}
    \caption{Hospital environment, no agents.}
    \label{fig:hospital_static}
\end{figure}

\begin{figure}[h!]
    \centering
    \includegraphics[width=0.99\linewidth]{figures/hospital_dynamic.png}
    \caption{Hospital environment, 15 agents.}
    \label{fig:hospital_dynamic}
\end{figure}
\end{comment}
% \newpage
\subsection{Experiment Four: Timing}
\label{sec:timing}
\begin{figure}[h!]
    % \vspace{-1cm}
    \subfloat[Timing for scan thread portrayed in Figure \ref{fig:info_flow}.]{\includegraphics[width=0.31\linewidth]{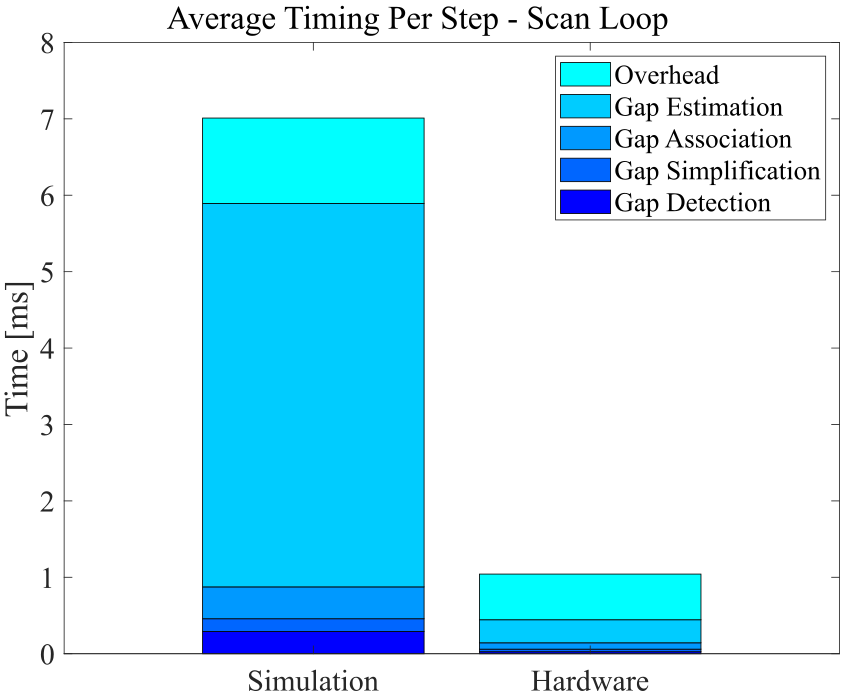}\label{fig:scan_loop_times}}
    \quad
    \subfloat[Timing for control thread portrayed in Figure \ref{fig:info_flow}.]{\includegraphics[width=0.31\linewidth]{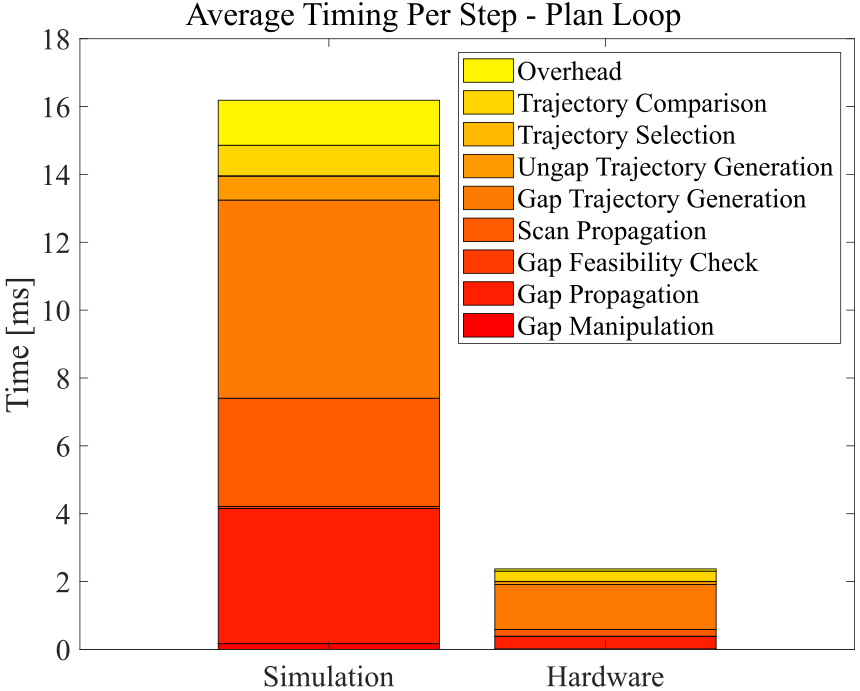} \label{fig:plan_loop_times}}
    \quad
    \subfloat[Timing for control thread portrayed in Figure \ref{fig:info_flow}.]{\includegraphics[width=0.31\linewidth]{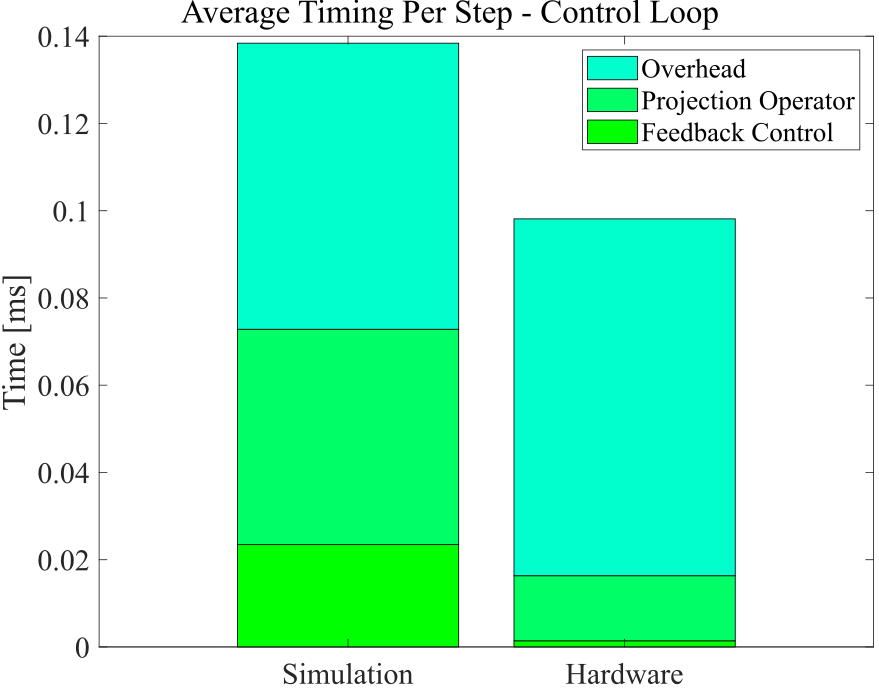} \label{fig:control_loop_times}}
    \caption{Timing breakdown for dynamic gap planner. The scan thread can run at $\approx 143$ Hz in simulation and $\approx 959$ Hz on hardware, but in practice it is set to the rate of the incoming sensor data in both simulation and hardware which is $25$ Hz. The planning thread can run at $\approx 62$ Hz (for an average of $5.76$ gaps per planning loop) in simulation and $\approx 337$ Hz (for an average of $5.80$ gaps per planning loop) on hardware, but in practice it is set to $5$ Hz within the \texttt{move\_base} framework. The control thread is bound to the planning thread.}
    % \vspace{-1cm}
\end{figure}

In this experiment, we report the average computation times in simulation and hardware for all baselines use during benchmarking. A more in-depth computational analysis is given for dynamic gap in Figures \ref{fig:scan_loop_times} - \ref{fig:control_loop_times}. All timing experiments in simulation are run in one scenario of the Factory environment with $15$ agents present.
% , and average overall computation times for other baselines are given in Table \ref{tab:benchmark_times}. 
\subsubsection{Dynamic Gap}
\subsubsection{Other Baselines}
Planning rates and implementation details for other baselines are shown in Table \ref{tab:benchmark_times}.

\newcommand \rowOneWidth{0.75cm}
\newcommand \rowTwoWidth{1.1cm}
\newcommand \rowThreeWidth{0.75cm}
\newcommand \rowFourWidth{0.75cm}
\newcommand \rowFiveWidth{1.0cm}
\newcommand \rowSixWidth{1.1cm}
\newcommand \rowSevenWidth{1.25cm}
\newcommand \rowEightWidth{0.75cm}
\newcommand \rowNineWidth{0.75cm}
\newcommand \rowTenWidth{0.75cm}
\newcommand \rowElevenWidth{1.0cm}

\begin{table}[h!]
    \centering
    \caption{Reported Computation Times (Hz) for Simulation Benchmarks}
    \begin{tabular}{ | >{\Centering}m{\rowOneWidth} | >{\Centering}m{\rowTwoWidth} |  >{\Centering}m{\rowThreeWidth} | >{\Centering}m{\rowFourWidth} |>{\Centering}m{\rowFiveWidth} | >{\Centering}m{\rowSixWidth} | >{\Centering}m{\rowSevenWidth} |  >{\Centering}m{\rowEightWidth} | >{\Centering}m{\rowNineWidth} | >{\Centering}m{\rowTenWidth} |  >{\Centering}m{\rowElevenWidth} |}
    \hline
    \textbf{TEB} & \textbf{CoHAN} & \textbf{MPC} & \textbf{DWA} & \textbf{RLCA} & \textbf{CADRL} & \textbf{DRL-VO} & \textbf{LfLH} & \textbf{PGap} & \textbf{DGap} & \textbf{DGap+} \\
    \hline
    71 & 201 & 39 & 6 & 593 & 539 & 36 & 22 & 26 & 47 & 62 \\
    \hline
    \end{tabular}
    \label{tab:benchmark_times}
\end{table}

\subsection{Experiment Five: Social Compliance Performance}
\label{sec:social}

% within the DGap$+$ cost function 
% We will refer to this
A relative velocity cost function based on \cite{teja_singamaneni_human-aware_2021} is added to improve the social compliance of the robot. This new planning variant is referred to as DGap$^+$Social.

\begin{equation} \label{eq:rel_vel_cost}
\mathrm{cost}_{\mbox{rel\_vel}}
  = \frac{%
      \Bigl(
        \max\bigl(\vec{V}_{\mathrm{rel}}\!\cdot\!\overrightarrow{P_rP_h},0\bigr)
        + \left\|\vec{V}_{r}\right\|
      \Bigr)%
    }{%
      \left\|\overrightarrow{P_rP_h}\right\|%
    }
\end{equation}

The first term involves a dot product in order to penalize paths that are directed towards a human. Moreover, the maximum term restricts the cost to situations in which the robot is approaching a human rather than moving away from them. The second term penalizes high velocities in the vicinity of a human.

\begin{figure}[b]
    \centering
    \includegraphics[width=0.85\linewidth]{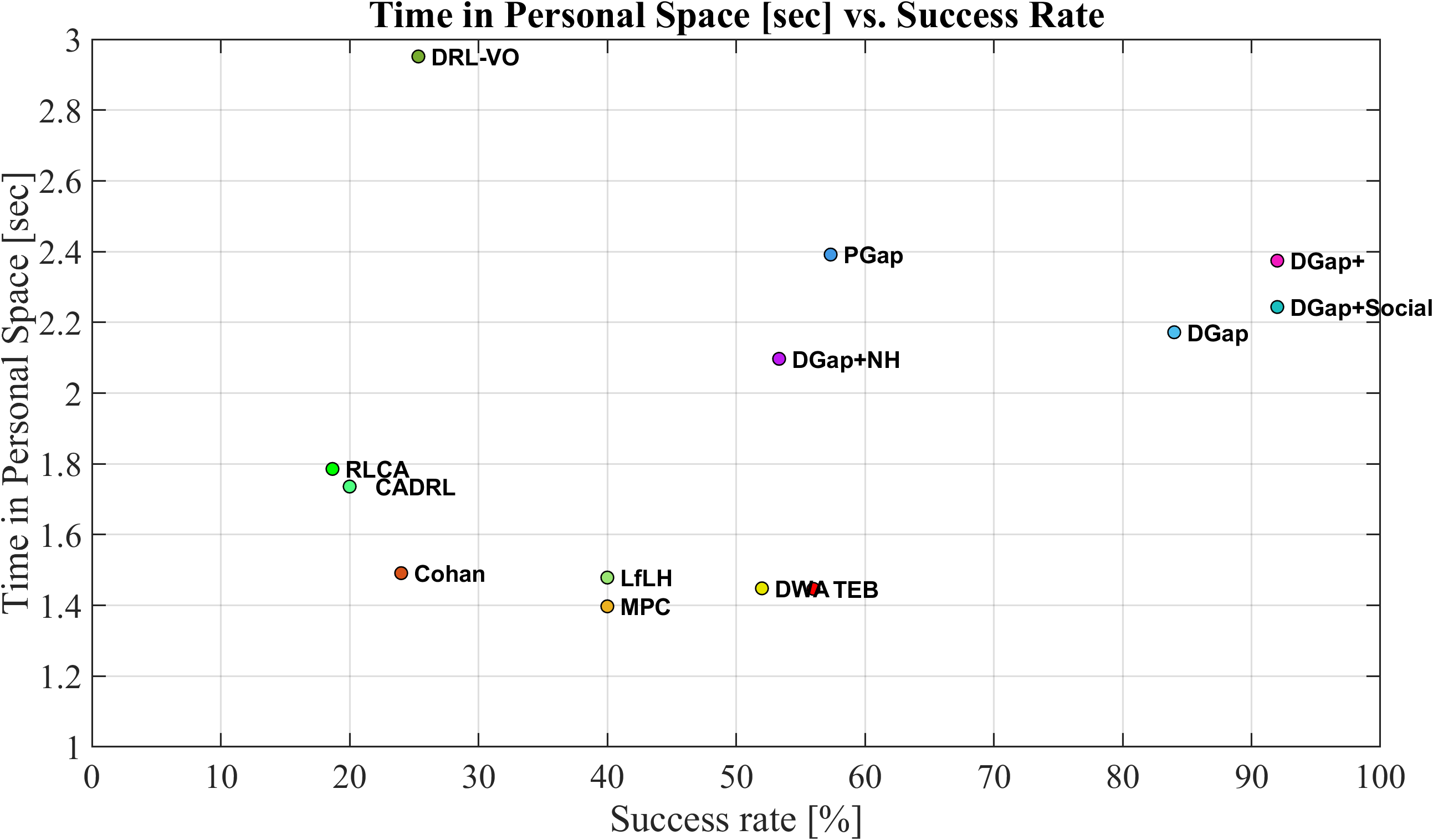}
    \caption{ Comparison of various navigation algorithms based on their time spent in personal space (in seconds) versus success rate (\%). }
    \label{fig:social_time_in_in_personal_and_success}
\end{figure}

Results related to social compliance were also collected from the same experiments previously described in Section \ref{sec:simulation_benchmarking}. As shown in Figure \ref{fig:social_time_in_in_personal_and_success}, DGap$^+$ and DGap$^+$Social achieved the highest success rates among all planners. However, they also exhibited relatively high time spent in personal space which is defined as a circle region surrounding the human with a radius of 0.5 m. This may be partially attributed to the fact that only successful trials were included in the plot. In densely crowded scenarios, many other planners likely failed early, producing no data for those challenging cases. In contrast, DGap methods successfully navigated through such environments which caused it to spend more time in close proximity to people, thereby increasing their average time in personal space.

Interestingly, classical planners such as MPC, DWA, and TEB showed both reasonably high success rates and comparable or lower time in personal space than most learned methods. This outcome is somewhat surprising, given that learning-based approaches are often claimed to yield superior social behavior. For instance, 90\% of DWA, TEB, and MPC trials had personal-space durations under 3 seconds, whereas 90\% of learned planners stayed under 4 seconds, excluding DRL-VO, which was below 7 seconds.
The CoHAN planner had one of the lower success rates. This was likely due to its tendency to attempt to pass in front of pedestrians, as well as its occasional collisions with the environment when turning corners.

% , which can be viewed as a variant of TEB with social navigation extensions,

\subsection{Experiment Six: Hardware Testing}
\label{sec:hardware_testing}

In this section, the dynamic gap planner is deployed on a TurtleBot2, a differential drive platform in which ROS code is executed from an onboard laptop. 
\subsubsection{Test One: Static obstacle course}

\begin{figure}[h!]
    \centering
    \includegraphics[width=0.99\linewidth]{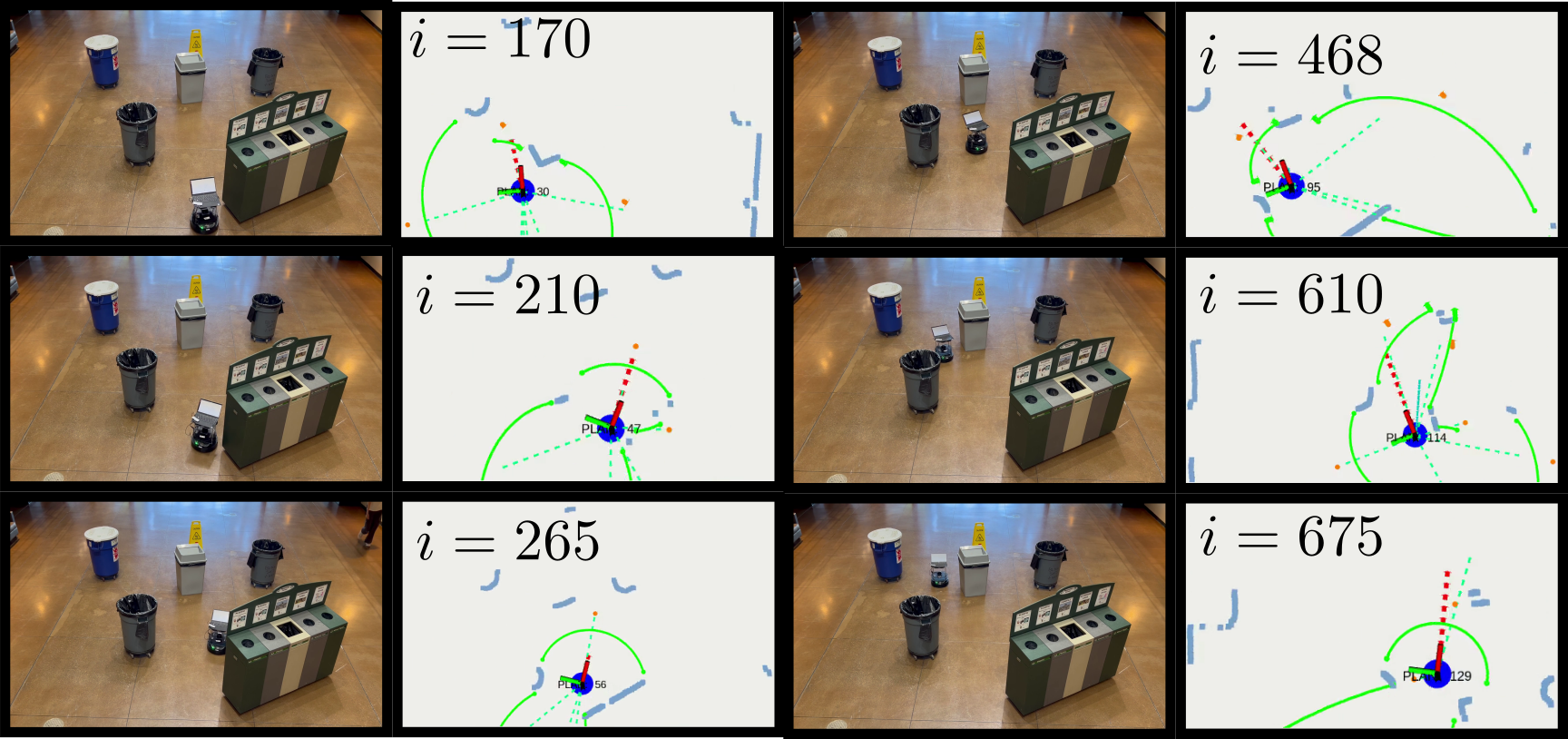}
    \caption{Hardware experiment one: static obstacle course. Left column depicts frames from the hardware setup and right column depicts online environment visualizations from the planner. The ego-robot (starting at the bottom of the frame) must travel to the top of the frame while avoiding the array of static obstacles.}
    \label{fig:hardware_test4}
\end{figure}

In this test, the ability of the dynamic gap planner to operate in a static setting is confirmed on hardware. The keyframes of this test are shown in Figure \ref{fig:hardware_test4}. In this scenario, the goal is placed at the top of the frame. At frame $i=210$, the ego-robot selects the gap in the center of the frame. By frame $i=468$, further gaps in the environment are detected. The planner selects the gap to the left of the frame and continues through this gap in frame $i=610$. By frame $i=675$, the planner has passed through this left gap and continues up the frame through another front-facing gap on its way to the goal.

% The planner begins at frame $i=210$ by entering the center of the environment. 

\subsubsection{Test Two: Translating Agent}

\begin{figure}[h!]
    \subfloat[Hardware experiment two: translating agent. Left column depicts frames from the hardware setup and right column depicts online environment visualizations from the planner. The ego-robot (starting at the bottom of the frame) must travel to a goal placed beyond the other TurtleBot (starting in the middle of the frame) which is translating back and forth.]{\includegraphics[width=0.49\linewidth]{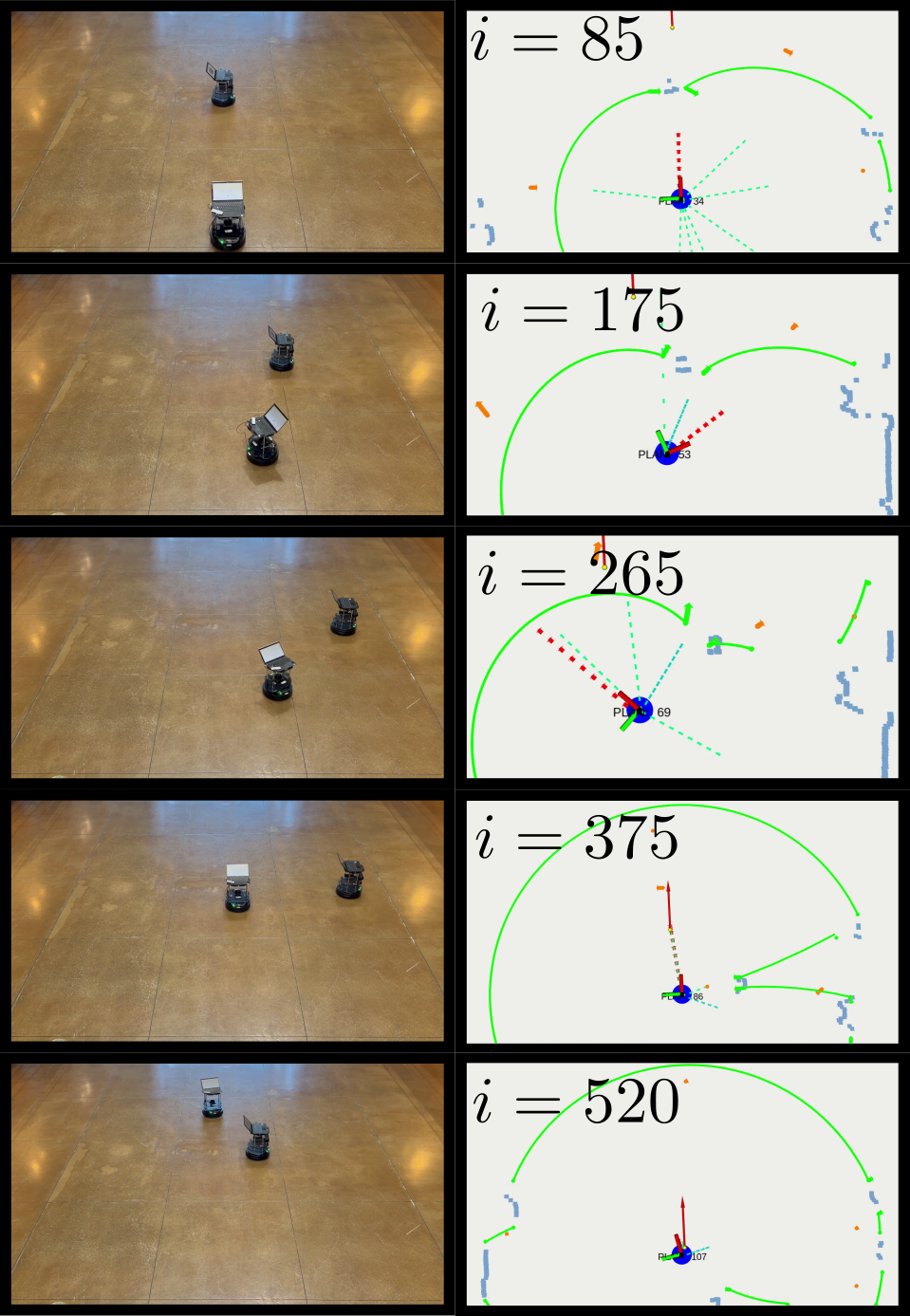}\label{fig:hardware_test1}}
    \quad
    \subfloat[Hardware experiment three: translating gap. Left column depicts frames from the hardware setup and right column depicts online environment visualizations from the planner. The ego-robot (starting at the bottom of the frame) must travel to a goal placed beyond the two TurtleBots that are translating back and forth so as to form a gap between them.]{\includegraphics[width=0.49\linewidth]{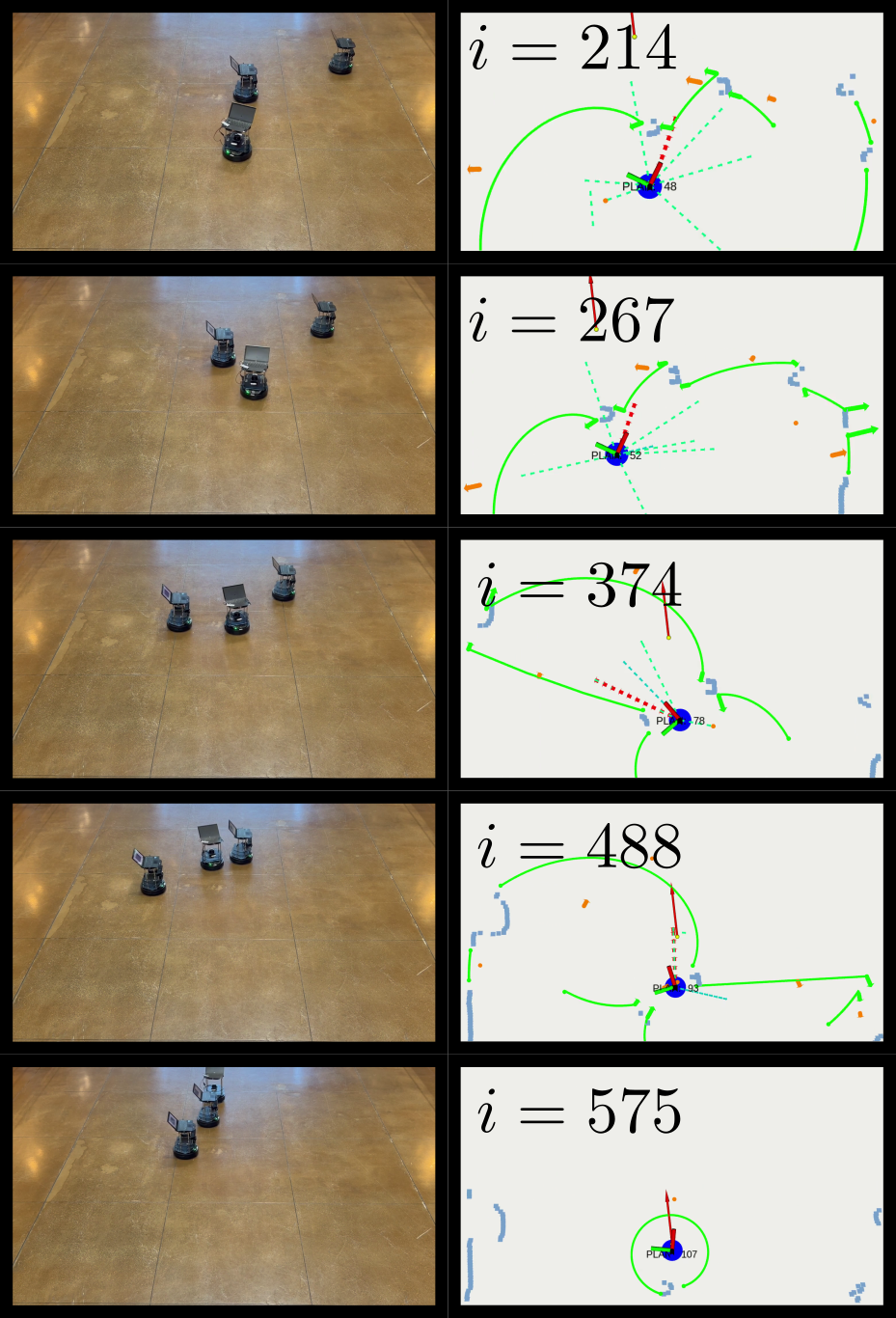}\label{fig:hardware_test2}}
    \caption{Hardware experiments two and three - translating scenarios.}
\end{figure}

For this test, a second TurtleBot2 platform is programmed to track a straight line path between two waypoints at a constant velocity. The ego-robot must travel past this agent en route to the goal. The keyframes of this test are shown in Figure \ref{fig:hardware_test1}.

\begin{comment}
\begin{figure}[h!]
    \centering
    \includegraphics[width=0.7\linewidth]{figures/hardware_cs1.png}
    \caption{Hardware experiment two: translating agent. Left column depicts frames from the hardware setup and right column depicts online environment visualizations from the planner. The ego-robot (starting at the bottom of the frame) must travel to a goal placed beyond the other TurtleBot (starting in the middle of the frame) which is translating back and forth.}
    \label{fig:hardware_test1}
\end{figure}
\end{comment}

At frame $i=85$, the ego-robot is tracking a previously generated trajectory through the middle of the frame towards the agent. At frame $i=175$, the agent completes the trajectory it had been tracking and selects a new trajectory that travels to the right of the frame. However, at frame $i=265$, predicts a collision with the agent and switches to plan to the left of the agent. At frame $i=375$, the agent reaches its right waypoint and begins to travel back to the left of the frame, but at this point the ego-robot has almost passed the agent. By frame $i=520$, the ego-robot has passed the agent and reached the goal.

\subsubsection{Test Three: Translating Gap}

In this test, two agents are programmed to track a straight line path between two waypoints at a constant velocity, creating a translating gap that the ego-robot can pass through en route to the goal. Keyframes from this test are shown in Figure \ref{fig:hardware_test2}.

\begin{comment}
\begin{figure}[h!]
    \centering
    \includegraphics[width=0.70\linewidth]{figures/hardware_cs2.png}
    \caption{Hardware experiment three: translating gap. Left column depicts frames from the hardware setup and right column depicts online environment visualizations from the planner. The ego-robot (starting at the bottom of the frame) must travel to a goal placed beyond the two TurtleBots that are translating back and forth so as to form a gap between them.}
    \label{fig:hardware_test2}
\end{figure}
\end{comment}

At frame $i=214$, the dynamic gap planner propagates the central gap forward in time and selects the trajectory that passes through this gap to be track. This trajectory continues to be tracked at frame $i=267$, and at frame $i=374$ this trajectory is completed. At this point, the planner generates trajectories through the central gap that are biased to the left of the gap. These trajectories are biased because of their propagated versions. At iteration $i=488$, the ego-robot switches to a trajectory that will reach the goal, and by $i=575$ the planner reaches the goal.

\subsubsection{Test Four: Receding Corridor}

\begin{figure}[h!]
    \centering
    \includegraphics[width=0.99\linewidth]{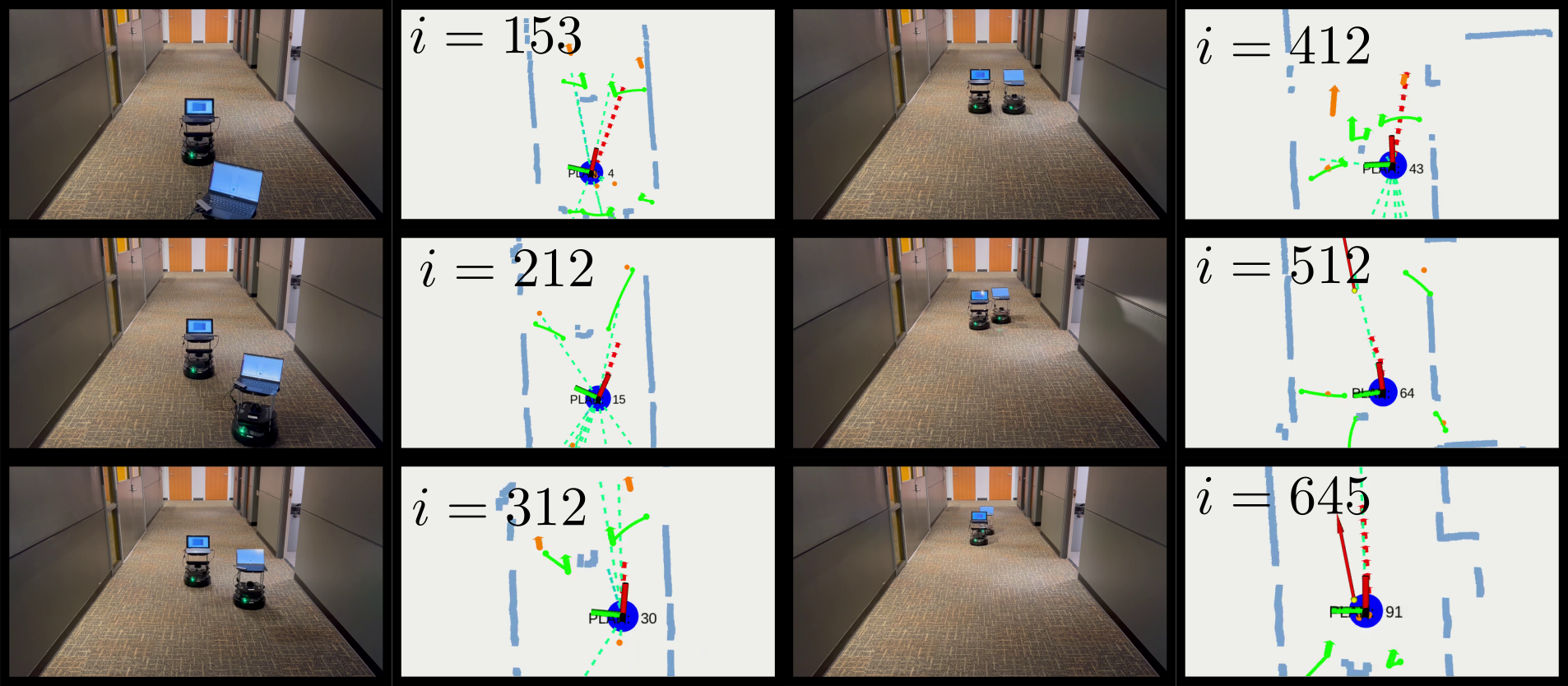}
    \caption{Hardware experiment four: corridor - receding. Left column depicts frames from the hardware setup and right column depicts online environment visualizations from the planner. The ego-robot (starting at the bottom of the frame) must travel to the end of the corridor while the agent in front of it also moves through the corridor.}
    \label{fig:hardware_test3}
\end{figure}

In this test, the planner must maneuver the ego-robot around an agent travelling down a corridor. Keyframes from this test are shown in Figure \ref{fig:hardware_test3}. In this scenario, the corridor is wide enough for the ego-robot to travel around the agent. The planner opts to do so in frame $i=212$ where the ego-robot begins to travel to the right of the agent as the planner select the gap to the right of the agent. During frames $i=312$ and $i=412$, the planner continues planning to the right of the agent. By frame $i=512$, the ego-robot begins to cut back in front of the agent in order to reach the goal which is in the center of the corridor. By frame $i=645$, the ego-robot reaches the goal in front of the agent.

\section{Conclusion} \label{sec:conclusion}
This chapter introduces a novel perception-informed gap-based planner known as dynamic gap. This planner is designed to operate in dynamic environments by tracking locally detected gaps of free space over time, propagating these gaps forward into the future to understand what subset of the free space remains feasible, and applying the PN geometric law from pursuit guidance theory to generate collision-free local trajectories under ideal conditions. Furthermore, the occupied polar regions referred to as ungaps are detected and planned within under circumstances in which no gaps exist. In this chapter, the dynamic gap planner is shown to outperform all other evaluated baselines in simulation benchmarks when tested on a holonomic robot platform. However, the planner performs worse when constrained to nonholonomic dynamics. This is largely due to the model mismatch between the holonomic model used to evaluate gap feasibility and generate local trajectories and the nonholonomic model that these trajectories are subsequently tracked on. In the future, the authors aim to integrate nonholonomic constraints directly into gap feasibility considerations and trajectory synthesis to mitigate this mismatch.

\bibliographystyle{IEEEtran}
\bibliography{max_references, abdel_references}

\end{document}